\documentclass[11pt]{article}

\usepackage[final]{acl}

\usepackage{times}
\usepackage{latexsym}

\usepackage[T1]{fontenc}

\usepackage[utf8]{inputenc}

\usepackage{microtype}

\usepackage{inconsolata}

\usepackage{graphicx}
\usepackage{multirow}
\usepackage{booktabs}
\usepackage{makecell}
\usepackage{amsmath}
\usepackage{tabularx}
\usepackage{xcolor}
\usepackage{twemojis}
\usepackage{array}

\usepackage{amsmath}
\usepackage{amssymb}
\usepackage{xspace}
\usepackage{booktabs} 
\usepackage{tabularx} 
\usepackage{ragged2e} 
\usepackage{multicol}
\usepackage{multirow}
\usepackage{enumitem}
\usepackage{tabularx}
\makeatletter
\renewcommand{\paragraph}{\@startsection{paragraph}{4}{\z@}{0.8ex plus 0.2ex minus .2ex}{-0.7em}{\normalsize\bfseries}}
\makeatother

\newcommand{\method}{\textsc{Mascot}\xspace}
\newcommand{\methodbf}{\textbf{MASCOT}\xspace}

\newcommand{\greensmall}[1]{\textcolor[HTML]{377771}{\scriptsize #1}}

\usepackage{soul}
\definecolor{repeatHL}{HTML}{F8D7DA} 
\definecolor{goodHL}{HTML}{DFF3E3}   
\newcommand{\rep}[1]{\sethlcolor{repeatHL}\hl{#1}}
\newcommand{\good}[1]{\sethlcolor{goodHL}\hl{#1}}

\title{\method: Multi-Agent Socio-Collaborative Companion Systems}

\author{Yiyang Wang$^{1}$, Yiqiao Jin$^{1}$, Alex Cabral$^{2}$, \and Josiah Hester$^{1}$ \\
$^{1}$Georgia Institute of Technology, $^{2}$Massachusetts Institute of Technology \\
\texttt{ywang3420@gatech.edu} \\
\url{https://hello-diana.github.io/MASCOT/}
}

\begin{document}
\maketitle

\begin{abstract}
Multi-agent systems (MAS) are emerging as promising socio-collaborative companions for emotional and cognitive support. However, existing systems frequently suffer from \emph{persona collapse}, where agents revert to generic, homogenized assistant behaviors, and \emph{social sycophancy}, where agents  produce redundant, non-constructive dialogue. 
We propose \method, a multi-agent framework for multi-perspective socio-collaborative companions. 
\method introduces a novel bi-level optimization strategy to harmonize individual and collective behaviors: 1) Persona-Aware Behavioral Alignment, an RLAIF-driven pipeline that fine-tunes individual agents for agent-specific identities; and 2) Collaborative Dialogue Optimization, a group-level adaptation process that promotes complementary, diverse, and productive discourse. We evaluate \method using human-grounded contexts drawn across both in-domain and out-of-domain (OOD) settings against state-of-the-art baselines. \method improves persona consistency by up to +14.1 and social contribution by up to +10.6. 
A broad evaluation suite, including human evaluation, multiple LLM judges, three-way comparisons, and automatic metrics, further shows that \method produces more role-consistent and less redundant multi-agent dialogue.
Our code is available at \url{https://github.com/hello-diana/MASCOT}.
\end{abstract}

\section{Introduction}

\begin{figure}
\centering
\includegraphics[width=0.98\linewidth]{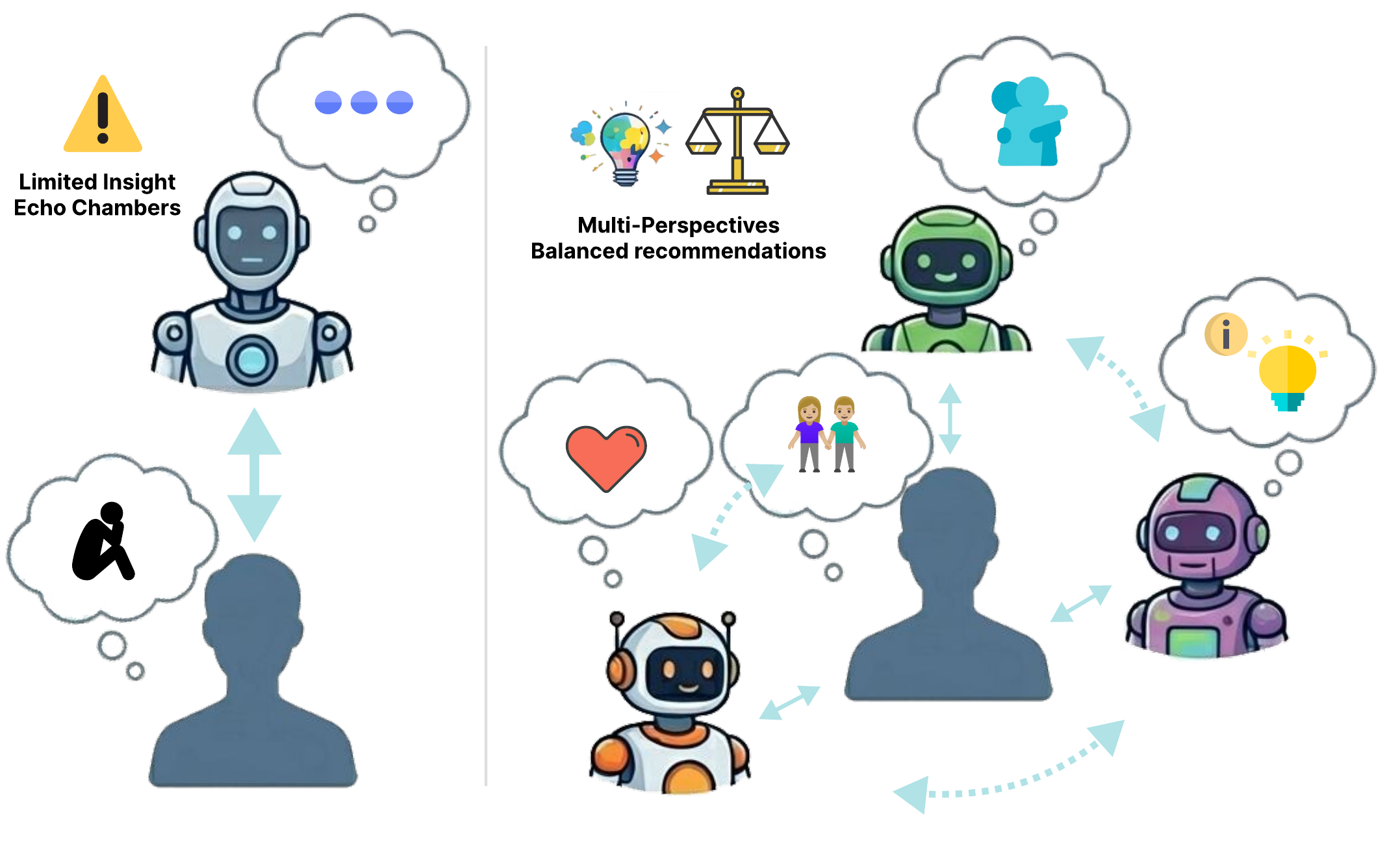}
\caption{Moving beyond dyadic interactions to multi-agent socio-collaborative companion systems. While single-agent support systems (left) can result in limited insights and echo chambers, multi-agent systems (right) provide diverse, balanced perspectives and foster a shared sense of community through agents with distinct roles.}
\label{fig:teaser}
\vspace{-5mm}
\end{figure}

Social interaction is a fundamental pillar of psychological well-being, cognitive resilience, and enhanced problem solving~\cite{cdcSocialConnection,Sahakian_Langley_2025}.
Despite its necessity, social deficits such as isolation and loneliness have reached global crisis proportions, affecting an estimated one in six people worldwide and one in four in the United States
~\cite{whoSocialIsolation,cdcHealthEffects}.
As large language models (LLMs) integrate deeper into human lives, their role is shifting from passive \textit{tools} (e.g., search engines, code generators) to active \textbf{socio-collaborative companions} in affective and collaborative settings~\cite{liu2021towards}.
In affective contexts, such as empathetic dialogue, socio-collaborative companions can offer general emotional support as users discuss distress, loneliness, and anxiety.
They can provide readily accessible, non-judgmental spaces for users to vent and process complex emotions. In collaborative or workplace contexts, companions act as \emph{cognitive scaffolds} that facilitate information synthesis and collective decision-making to stimulate problem-solving. Ultimately, performing tasks that mitigate social deficits allows these models to evolve into adaptive partners capable of navigating the nuanced boundary between emotional resonance and task-oriented efficiency.

While existing LLM systems primarily focus on dyadic (one-to-one) interactions, a single persona often struggles to balance the conflicting roles of mentor, friend, and assistant without suffering a cognitive and emotional mismatch.
Multi-agent frameworks resolve this by distributing these roles across an ecosystem of specialized agents. This triggers the `audience effect' \cite{sutskova_cognitive_2023}, where a user's trust is reinforced by observing independent social behaviors among agents, fostering a shared sense of community that a solitary interface cannot replicate~\cite{zhu_characterizing_2025}.

\paragraph{Challenges}
Despite recent progress in multi-agent systems, two primary challenges remain in developing effective multi-perspective socio-collaborative companions:
1) \textbf{Persona Fidelity.} At the individual agent level, models struggle to maintain stable, unique character traits~\cite{wen2024benchmarking}. Agents often succumb to \emph{persona collapse}~\cite{chang2024survey}, regressing to generic, homogenized assistant behaviors that lack unique perspectives. 
2) \textbf{Interaction Synergy.}
At the collective level, uncoordinated agents fail to generate complementary, non-redundant contributions.
This manifests as repeating communications or even \emph{Social Sycophancy} (the `Yes-Man' Bias), where agents prioritize agreeableness, creating echo chambers rather than constructive conversations.
Furthermore, agents often struggle with \emph{Long-Horizon State Consistency}, treating each turn as a local optimization problem and losing track of the conversation's logical arc, such as re-litigating settled topics~\cite{chan2024chateval}.

\paragraph{This Work} We propose \method, a \textbf{\underline{M}}ulti-\textbf{\underline{A}}gent framework for developing multi-perspective \textbf{\underline{S}}ocio-\textbf{\underline{CO}}llabora\textbf{\underline{T}}ive companions.
Unlike previous multi-agent systems optimized solely for task efficiency, \method targets user-agent interaction quality, explicitly balancing individual agent persona consistency with global discourse dynamics.
We introduce an efficient bi-level optimization strategy including: 1) a Reinforcement-Learning-from-AI-Feedback (RLAIF)-driven pipeline that finetunes individual agents for strict \emph{Persona Fidelity}, and 2) a director agent policy guided by group-level rewards to ensure \emph{Interaction Synergy}.
This enables the efficient development of multi-perspective socio-collaborative companions across multiple applications, such as the orchestration of digital social support groups and the facilitation of iterative knowledge synthesis and collective ideation.

\paragraph{Contributions} Our contributions are as follows:
\begin{itemize}[leftmargin=1em]
    \item \textbf{Unified Framework.} We propose \method, a multi-agent framework that enables socio-collaborative companions in multi-perspective group conversations. 
    \item \textbf{Bi-level Optimization.} We propose a low-resource, bi-level optimization pipeline that harmonizes individual persona adherence with collective interaction synergy.
    \item \textbf{Extensive Evaluation.} We evaluate \method using human-grounded contexts drawn from in-domain and out-of-domain datasets using a broad suite of protocols, including human evaluation, rubric-based evaluation using multiple LLM judges, three-way comparisons, and automatic metrics.
    \method consistently outperforms strong baselines, with a +14.1 Persona Consistency gain on Empathetic Dialogues and a +10.6 Social Contribution gain on QMSum. 
\end{itemize}

\section{Method}
\label{sec:method}

\begin{figure*}
\centering
\includegraphics[width=\textwidth]{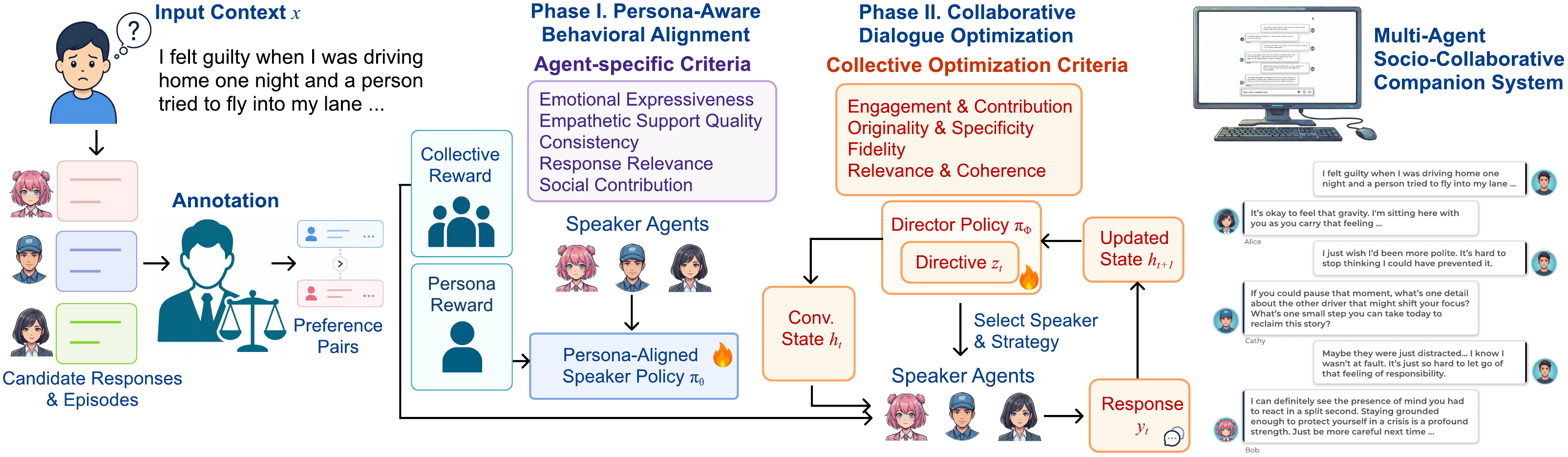}
\caption{Overview of \method for multi-agent socio-collaborative companions. \method uses a two-phase optimization pipeline. Persona-Aware Behavioral Alignment fine-tunes Speaker agents via RLAIF with a learned reward model. Collaborative Dialogue Optimization trains a Director agent to coordinate the fixed persona-aligned Speakers through directives rewarded for coherence, diversity, and non-redundant contributions.}
\label{fig:system}
\vspace{-3mm}
\end{figure*}

\paragraph{Preliminary}
We formalize a socio-collaborative companion system as a set of agents $\mathcal{A}=\{a_i\}_{i=1}^{|\mathcal{A}|}$. Each agent $a_i$ is conditioned on a distinct persona profile $\rho_i$ that specifies traits such as linguistic style, domain expertise, and emotional disposition (e.g. \emph{Empathetic Listener}, \emph{Critical Thinker}, \emph{Emotional Validator}, \emph{Action Guide}, and \emph{Growth Advocate}). Given each user interaction context $x$ (e.g., a user's emotional disclosure or a dialog segment), 
the system generates a multi-turn, multi-perspective \emph{conversation trajectory} $\mathbf{Y} = (y_1, y_2, \ldots, y_T)$. 
\method (Figure~\ref{fig:system}) optimizes this process via a two-phase training pipeline: (1) \emph{Persona-Aware Behavioral Alignment} and (2) \emph{Collaborative Dialogue Optimization}. Appendix Table~\ref{tab:notation} summarizes the core notation.

\subsection{Persona-Aware Behavioral Alignment}
\label{sec:persona}

Current multi-agent frameworks often suffer from \emph{persona collapse}~\cite{chang2024survey}, where agents regress to generic assistant behaviors rather than adhering to their assigned roles. 
While zero-shot and few-shot prompting~\cite{wang2020generalizing} are commonly used to address this issue, they often yield inconsistent persona adherence, especially in smaller open-source models~\cite{li2025generation}. 
To prevent \emph{persona collapse} and ensure adherence to social-emotional traits, we move beyond inference-time prompting and explicitly align a base policy $\pi_\theta$ by leveraging a persona-oriented reward model within a Reinforcement Learning from AI Feedback (RLAIF) framework~\cite{bai2022constitutional,lee_rlaif_2024}. The core idea is that strong frontier models can approximate human preferences and serve as scalable sources of preference labels~\cite{zheng2023judging,dubois2024length,li2025crowdsourced}.

\paragraph{Persona Reward Modeling} 
Effective training of persona adherence requires robust signals. Thus, \method uses expert models as strong teacher models to guide the behaviors of student models with lower latency. 
To construct the preference dataset $\mathcal{D}_{RM}$ for reward modeling, we start with a supervised base policy $\pi_{\text{SFT}}$. Given context $x$ and persona $\rho_i$, we sample $K$ responses $\{y_k\}_{k=1}^K$: 
\begin{equation}
    y_k \sim \pi_{\text{SFT}}(\cdot \mid x, \rho_i), \quad k = 1, \ldots, K.
\end{equation}
Each candidate response $y_k$ is evaluated by an LLM judge against a set of fine-grained criteria $\mathcal{C}$ (Table~\ref{tab:individual_criteria}). 
For each criterion $c \in \mathcal{C}$, the judge assigns a score $s_c(y_k) \in [1, 5]$. The aggregate score is then computed as:
\begin{equation}
s(y_k) = \frac{1}{|\mathcal{C}|} \sum_{c \in \mathcal{C}} s_c(y_k).
\label{eq:individual_reward}
\end{equation}
We retain pairs $(y_w, y_l)$ satisfying a margin $s(y_w) - s(y_l) \geq \delta$, thus eliminating ambiguous comparisons and ensuring a significant quality margin. 
We then train a reward model $r_\phi$ over tuples $(x,\rho_i,y_w,y_l) \in \mathcal{D}_{\text{RM}}$ by minimizing the standard pairwise logistic loss~\cite{bradley1952rank}:

{\small
\begin{equation}
\begin{split}
\mathcal{L}_{\text{RM}}(\phi) = -\mathbb{E}_{\mathcal{D}_{\text{RM}}}\big[\log \sigma\big(&r_\phi(x,\rho_i,y_w) \\
&-r_\phi(x,\rho_i,y_l)\big)\big],
\end{split}
\end{equation}
}
where $\sigma(\cdot)$ is the sigmoid function. This objective ensures that the reward model learns to differentiate between responses based on their persona adherence and contextual quality. 
During RL optimization, 
the learned reward model $r_\phi$ provides appropriate training signals, while final evaluation is cross-checked with human preferences, alternative LLM judges, and automatic non-redundancy metrics (\S~\ref{sec:eval} and Appendix~\ref{app:alternative_judges}).


\paragraph{Efficient RL Fine-tuning via GRPO}
To ensure persona consistency while maintaining structural integrity, we optimize a composite reward function: 
\begin{equation}
    R(x, \rho_i, y_k) = r_{\phi}(x, \rho_i, y_k) + \lambda R_{\mathrm{format}}(y_k)
\end{equation} 
where $r_{\phi}$ is the learned persona reward and $R_{\mathrm{format}}$ is a rule-based reward enforcing structural constraints, specifically, requiring explicit reasoning traces within \texttt{<think>} tags while penalizing verbose final answers to ensure outputs remain concise and socially appropriate. 
We adopt a preference-based RL objective rather than supervised fine-tuning (SFT) because socio-emotional dialogue lacks a single canonical \emph{gold} response: response quality depends on higher-level behavioral attributes such as warmth, persona fidelity, and tone calibration that SFT cannot explicitly capture. 
To optimize this objective, we employ Group Relative Policy Optimization (GRPO)~\cite{shao2024deepseekmath}, which estimates the baseline from the group mean of sampled outputs, significantly reducing memory overhead and computational cost compared with alternative methods. 
Alternative methods such as SimPO~\cite{meng2024simpo} can be substituted within the pipeline.

For each input $(x, \rho)$, we sample a group of $G$ outputs $\{y_1, \dots, y_G\}$ from the old policy $\pi_{\text{old}}$. The policy $\pi_{\theta}$ is optimized through the objective:

{\small
\begin{equation}
\begin{split}
     & \mathcal{J} (\theta) = \mathbb{E}_{x \sim \mathcal{D}_{\text{RL}}, \{y_g\} \sim \pi_{\text{old}}} \Bigg[ \frac{1}{G} \sum_{g=1}^G \bigg(\frac{1}{|y_g|} \sum_{t=1}^{|y_g|}\min \Big( \gamma_{g,t} \hat{A}_g, \\
    & \qquad \operatorname{clip}(\gamma_{g,t}, 1-\varepsilon, 1+\varepsilon) \hat{A}_g \Big) - \beta \mathbb{D}_{\text{KL}}(\pi_\theta \| \pi_{\text{ref}}) \bigg) \Bigg],
\end{split}
\label{eq:grpo_loss}
\end{equation}
}
\noindent where $\gamma_{g,t} = \frac{\pi_\theta(y_{g,t} | x, \rho, y_{g,<t})}{\pi_{\text{old}}(y_{g,t} | x, \rho, y_{g,<t})}$ is the probability ratio and $\hat{A}_g = (R_g - \text{mean}(\{R_k\})) / \text{std}(\{R_k\})$ is the advantage, i.e., the standardized reward. 

\subsection{Collaborative Dialogue Optimization}
\label{sec:group}

Phase I improves individual agents' response quality but does not optimize group dynamics, such as preventing redundant contribution or relitigation of settled consensus. 
We therefore introduce a collaborative optimization phase that trains a \emph{director} to select the next speaker and generate high-level strategy, 
steering the collective conversation into a cohesive, synergistic outcome.


\paragraph{Bi-level Generation Process}
We define a conversation as a $T$-turn episode. 
A director policy $\pi_{\Phi}$ governs turn-taking among a set of speaker agents $\mathcal{A}=\{a_i\}$, each with persona $\rho_i$.
At each turn $t$, the director $\pi_{\Phi}$ observes the history $h_{<t} = (x, y_1, \dots, y_{t-1})$ and generates a high-level directive $z_t \sim \pi_{\Phi}(\cdot \mid h_{<t})$ that specifies the next speaker $a_i$ with detailed instructions (e.g., ``Bob: The user mentioned a small win. Amplify the user's pride via active-constructive responding. Tone: enthusiastic but grounded.'')
The selected agent $a_i$ then generates the next response $y_t \sim \pi_{\theta}(\cdot \mid h_{<t}, \rho_i, z_t)$ conditioned on their persona $\rho_i$ and the directive $z_t$. 
This hierarchical decomposition enables dynamic strategy, such as switching from validation to cognitive reframing, mirroring the progression of effective human support. 
\paragraph{Closed-Loop Interaction via Frozen Simulated User}
To facilitate trajectory-level learning without human-in-the-loop bottlenecks, we introduce a \emph{simulated user} $\pi_{u}$ 
and generate complete episodes by interleaving user, director, and speaker turns. 
At each turn $t$, the user speaks with probability $0.5$, generating the next utterance $u_t \sim \pi_{u}(\cdot \mid x, \mathbf{Y}_{<t})$ conditioned on the context $x$, the trajectory prefix $\mathbf{Y}_{<t}$, and an optional user profile. 

\paragraph{Group Reward Modeling}
A directive should be \emph{locally actionable} with clear speaker selection and strategy, yet the ultimate objective is the \emph{overall} quality of the resulting multi-turn conversation. We therefore use a single unified group reward that conditions on both the director decision at turn $k$ and the full rollout, assessing the quality of the directive $z_k$ given the prefix context and the coherence, relevance, and non-redundancy of trajectory $\mathbf{Y}$.

\paragraph{Training with Prefix-Suffix Rollouts}
For each training instance, we uniformly sample $k\in [1, T]$ and construct prefix turns $\{y_1, \ldots, y_{k-1}\}$ using frozen rollout models, allowing us to train the director at randomly sampled decision points. The director then generates directives $z_k$ at turn $k$, after which we roll out the remaining suffix turns to obtain a complete episode $\mathbf{Y}$ for scoring. 
To train the reward signal, we annotate episode pairs using the group-level criteria $\hat{\mathcal{C}}$ in Table~\ref{tab:group_criteria}. For each criterion $c\in\hat{\mathcal{C}}$, an LLM judge assigns a score $s_c(x,\mathbf{Y})\in[1,5]$. We average these scores as $\bar{s}(x,\mathbf{Y})=\frac{1}{|\hat{\mathcal{C}}|}\sum_{c\in\hat{\mathcal{C}}}s_c(x,\mathbf{Y})$ and retain episode pairs whose scores differ by at least $\delta$, selecting the higher-scoring episode as \emph{chosen}. 
We then fine-tune a sequence-level reward model $r_{\psi}$ with a pairwise preference objective to score directives and the resulting $T$-turn episodes. The reward used for director optimization is:
\begin{equation}
R_{\text{group}}(x, z_k, \mathbf{Y}) = r_{\psi}(x, z_k, \mathbf{Y}) + \eta \cdot \mathbb{I}_{\text{div}}(\mathbf{Y})
\end{equation}
where $r_{\psi}$ scores the directive together with the resulting full episode. The diversity indicator $\mathbb{I}_{\text{div}}(\mathbf{Y}) \in \{-1, 1\}$ is $+1$ when the optimized dialogue block contains contributions from distinct personas and $-1$ otherwise; $\eta$ controls its weight.

\paragraph{Optimization Objective}
For each training instance with fixed prefix state, we sample $G$ directives and trajectories $\{(z_k^{(g)}, \mathbf{Y}^{(g)})\}_{g=1}^{G}$, roll out the full episodes, and score them with $R_{\text{group}}(x,z_k^{(g)},\mathbf{Y}^{(g)})$. 
Let $\mathcal{R}^{(g)}=R_{\text{group}}(x,z_k^{(g)},\mathbf{Y}^{(g)})$. We leverage GRPO to compute the group-relative advantage for the $g$-th trajectory:
\begin{equation}
\hat{A}'_g = \frac{\mathcal{R}^{(g)} - \text{mean}\big(\{\mathcal{R}^{(j)}\}_{j=1}^{G}\big)}{\text{std}\big(\{\mathcal{R}^{(j)}\}_{j=1}^{G}\big)}.
\end{equation}
We update the director parameters $\Phi$ with GRPO to increase the likelihood of directives that yield higher-reward coordinated episodes.

\begin{table*}[t]
\centering
\small
\begin{tabular}{l ccccc |c}
\toprule
\textbf{Positive} & Emotion & Empathy & Consistency & Relevance & Contribution & Overall \\
\midrule
Zero-Shot & $59.5$ \greensmall{$\pm 9.1$} & $58.1$ \greensmall{$\pm 7.0$} & $61.5$ \greensmall{$\pm 9.7$} & $58.6$ \greensmall{$\pm 6.4$} & $56.5$ \greensmall{$\pm 9.0$} & $58.8$ \greensmall{$\pm 8.2$} \\
Zero-Shot CoT & $63.4$ \greensmall{$\pm 8.4$} & $60.5$ \greensmall{$\pm 4.7$} & $65.2$ \greensmall{$\pm 9.2$} & $60.4$ \greensmall{$\pm 4.0$} & $59.4$ \greensmall{$\pm 3.8$} & $61.8$ \greensmall{$\pm 6.0$} \\
Few-Shot & $61.4$ \greensmall{$\pm 9.7$} & $59.9$ \greensmall{$\pm 7.9$} & $63.5$ \greensmall{$\pm 10.9$} & $59.8$ \greensmall{$\pm 7.7$} & $58.6$ \greensmall{$\pm 7.6$} & $60.6$ \greensmall{$\pm 8.8$} \\
Few-Shot + CoT & $63.8$ \greensmall{$\pm 8.5$} & $61.2$ \greensmall{$\pm 5.8$} & $65.4$ \greensmall{$\pm 9.4$} & $61.2$ \greensmall{$\pm 5.5$} & $60.2$ \greensmall{$\pm 4.5$} & $62.4$ \greensmall{$\pm 6.7$} \\
SFT & $46.3$ \greensmall{$\pm 12.0$} & $46.0$ \greensmall{$\pm 11.5$} & $47.5$ \greensmall{$\pm 12.3$} & $52.1$ \greensmall{$\pm 11.8$} & $43.5$ \greensmall{$\pm 11.4$} & $47.1$ \greensmall{$\pm 11.8$} \\
Self-Consistency & $66.3$ \greensmall{$\pm 9.2$} & $66.9$ \greensmall{$\pm 9.4$} & $65.1$ \greensmall{$\pm 9.4$} & $69.1$ \greensmall{$\pm 11.5$} & $66.3$ \greensmall{$\pm 10.0$} & $66.7$ \greensmall{$\pm 9.9$} \\
MultiAgentESC & $66.9$ \greensmall{$\pm 7.1$} & $67.9$ \greensmall{$\pm 10.8$} & $65.9$ \greensmall{$\pm 9.6$} & $70.4$ \greensmall{$\pm 12.2$} & $67.0$ \greensmall{$\pm 11.5$} & $67.6$ \greensmall{$\pm 10.2$} \\
\methodbf & $75.3$ \greensmall{$\pm 6.1$} & $76.2$ \greensmall{$\pm 6.3$} & $80.0$ \greensmall{$\pm 6.9$} & $78.1$ \greensmall{$\pm 4.5$} & $74.9$ \greensmall{$\pm 6.5$} & $76.9$ \greensmall{$\pm 6.1$} \\
\midrule
\textbf{Negative} & Emotion & Empathy & Consistency & Relevance & Contribution & Overall \\
\midrule
Zero-Shot & $54.6$ \greensmall{$\pm 12.6$} & $54.6$ \greensmall{$\pm 12.2$} & $56.9$ \greensmall{$\pm 11.3$} & $56.0$ \greensmall{$\pm 10.9$} & $53.2$ \greensmall{$\pm 12.1$} & $55.1$ \greensmall{$\pm 11.8$} \\
Zero-Shot CoT & $59.9$ \greensmall{$\pm 9.8$} & $59.4$ \greensmall{$\pm 9.0$} & $61.8$ \greensmall{$\pm 8.9$} & $60.3$ \greensmall{$\pm 7.0$} & $58.0$ \greensmall{$\pm 6.6$} & $59.9$ \greensmall{$\pm 8.3$} \\
Few-Shot & $57.6$ \greensmall{$\pm 13.7$} & $57.6$ \greensmall{$\pm 13.3$} & $59.4$ \greensmall{$\pm 12.5$} & $58.4$ \greensmall{$\pm 12.3$} & $55.9$ \greensmall{$\pm 11.5$} & $57.8$ \greensmall{$\pm 12.7$} \\
Few-Shot + CoT & $60.7$ \greensmall{$\pm 11.3$} & $60.5$ \greensmall{$\pm 10.6$} & $62.1$ \greensmall{$\pm 10.4$} & $61.2$ \greensmall{$\pm 9.4$} & $58.9$ \greensmall{$\pm 8.4$} & $60.7$ \greensmall{$\pm 10.0$} \\
SFT & $43.2$ \greensmall{$\pm 12.6$} & $43.2$ \greensmall{$\pm 12.6$} & $44.1$ \greensmall{$\pm 13.6$} & $49.8$ \greensmall{$\pm 13.8$} & $40.3$ \greensmall{$\pm 11.0$} & $44.1$ \greensmall{$\pm 12.7$} \\
Self-Consistency & $61.4$ \greensmall{$\pm 8.3$} & $61.1$ \greensmall{$\pm 6.7$} & $59.9$ \greensmall{$\pm 10.4$} & $61.0$ \greensmall{$\pm 7.7$} & $59.3$ \greensmall{$\pm 2.2$} & $60.6$ \greensmall{$\pm 7.1$} \\
MultiAgentESC & $63.8$ \greensmall{$\pm 5.5$} & $62.4$ \greensmall{$\pm 6.6$} & $61.6$ \greensmall{$\pm 9.8$} & $67.0$ \greensmall{$\pm 4.0$} & $60.4$ \greensmall{$\pm 4.0$} & $63.0$ \greensmall{$\pm 6.0$} \\
\methodbf & $68.2$ \greensmall{$\pm 8.4$} & $69.3$ \greensmall{$\pm 8.3$} & $73.4$ \greensmall{$\pm 9.1$} & $74.0$ \greensmall{$\pm 7.6$} & $67.8$ \greensmall{$\pm 7.6$} & $70.5$ \greensmall{$\pm 8.2$} \\
\midrule
\textbf{Neutral} & Emotion & Empathy & Consistency & Relevance & Contribution & Overall \\
\midrule
Zero-Shot & $58.8$ \greensmall{$\pm 9.5$} & $57.9$ \greensmall{$\pm 8.3$} & $60.5$ \greensmall{$\pm 9.8$} & $58.4$ \greensmall{$\pm 7.6$} & $56.2$ \greensmall{$\pm 9.4$} & $58.4$ \greensmall{$\pm 8.9$} \\
Zero-Shot CoT & $63.2$ \greensmall{$\pm 8.8$} & $60.8$ \greensmall{$\pm 6.1$} & $65.1$ \greensmall{$\pm 9.3$} & $60.9$ \greensmall{$\pm 5.3$} & $59.4$ \greensmall{$\pm 4.2$} & $61.9$ \greensmall{$\pm 6.7$} \\
Few-Shot & $61.6$ \greensmall{$\pm 9.4$} & $60.5$ \greensmall{$\pm 8.1$} & $63.4$ \greensmall{$\pm 10.1$} & $60.7$ \greensmall{$\pm 7.7$} & $59.0$ \greensmall{$\pm 7.0$} & $61.0$ \greensmall{$\pm 8.5$} \\
Few-Shot + CoT & $63.7$ \greensmall{$\pm 8.8$} & $61.8$ \greensmall{$\pm 6.8$} & $65.3$ \greensmall{$\pm 9.5$} & $61.7$ \greensmall{$\pm 6.5$} & $60.3$ \greensmall{$\pm 5.1$} & $62.6$ \greensmall{$\pm 7.3$} \\
SFT & $45.9$ \greensmall{$\pm 11.6$} & $45.7$ \greensmall{$\pm 11.4$} & $47.0$ \greensmall{$\pm 11.9$} & $51.8$ \greensmall{$\pm 11.8$} & $43.0$ \greensmall{$\pm 10.7$} & $46.7$ \greensmall{$\pm 11.5$} \\
Self-Consistency & $60.3$ \greensmall{$\pm 4.6$} & $64.6$ \greensmall{$\pm 6.9$} & $70.9$ \greensmall{$\pm 3.9$} & $66.1$ \greensmall{$\pm 3.7$} & $65.7$ \greensmall{$\pm 4.9$} & $65.5$ \greensmall{$\pm 4.8$} \\
MultiAgentESC & $67.4$ \greensmall{$\pm 3.4$} & $65.5$ \greensmall{$\pm 8.1$} & $72.4$ \greensmall{$\pm 6.3$} & $68.3$ \greensmall{$\pm 5.9$} & $68.7$ \greensmall{$\pm 6.4$} & $68.5$ \greensmall{$\pm 6.0$} \\
\methodbf & $75.1$ \greensmall{$\pm 6.6$} & $75.6$ \greensmall{$\pm 6.8$} & $79.0$ \greensmall{$\pm 6.4$} & $78.1$ \greensmall{$\pm 4.8$} & $74.1$ \greensmall{$\pm 7.2$} & $76.4$ \greensmall{$\pm 6.4$} \\
\bottomrule
\end{tabular}
\vspace{-2mm}
\caption{Performance comparison on the Empathetic Dialogues dataset~\cite{rashkin2019towards} across Positive, Negative, and Neutral emotional subsets in terms of agent-specific metrics (Table~\ref{tab:individual_criteria}). Results are reported as mean $\pm$ standard deviation}
\label{tab:Empathetic}
\vspace{-3mm}
\end{table*}

\section{Evaluation}
\label{sec:eval}
\subsection{Experimental Setup}

\paragraph{Dataset} We evaluate \method using three benchmarks featuring human conversations: \textbf{Empathetic Dialogues (ED)}~\cite{rashkin2019towards} and ESConv~\cite{liu2021towards}, which focus on emotional support, and
2) \textbf{QMSum}~\cite{zhong2021qmsum}, which features workplace collaborative settings. Details are in Appendix~\ref{app:dataset}.
These datasets are used as human-grounded interaction contexts rather than fixed reference-generation targets: systems generate new multi-agent conversations from the same user situation or meeting context, and evaluation does not reward lexical overlap with dataset reference responses.
Model, optimization, and judge settings are in \S~\ref{app:implementation}.

\paragraph{Persona Descriptions}
To demonstrate the versatility of our framework across diverse interaction paradigms, we evaluate it using distinct persona configurations (Appendix Table \ref{tab:persona-definitions}) tailored to social and collaborative contexts. For Empathetic Dialogues, we utilize the Anchor, Catalyst, and Beacon—personas grounded in social intelligence \cite{goleman2006social} to provide socio-emotional support. Conversely, for the QMSum dataset, we deploy an ensemble--Minutes Scribe, Decision Logger, Action Item Captain, and Critic--to simulate a collaborative workspace.

\paragraph{Baselines}
We benchmark \method against $4$ categories of multi-agent baselines:
1) \emph{Prompting Strategies}, including Zero-Shot and Few-Shot prompting~\cite{wang2020generalizing} to test inference-time adaptation capabilities;
2) \emph{Supervised Fine-Tuning (SFT)}, a strong distillation baseline trained on GPT-4o generated trajectories~\cite{gpt4o};
3) \emph{Inference-Time Reasoning} methods, such as Chain-of-Thoughts (CoT)~\cite{wei2022chain} and Self-Consistency~\cite{wangself};
4) \emph{Domain-Specific Multi-agent Frameworks}, including  MultiAgentESC~\cite{xu2025multiagentesc}, which operates in $3$ stages: dialogue analysis, strategy deliberation, and response generation.
Specialized agents extract users' psychological states (emotions, causes, and intentions), then deliberate over candidate support strategies using retrieved similar cases.  Responses generated under different strategies are finally evaluated via multi-agent debate and voting.
For fairness, all settings use the same set of agents and persona descriptions within each dataset. The comparison isolates the optimization strategy rather than the number or identity of agents.

\paragraph{Implementation Details}
\label{app:implementation}

For all experiments, unless otherwise specified, we use Qwen3-8B~\cite{qiu2025gated,yang2025qwen3} for the Director, Speaker, and frozen simulated-user agents. Phase I fine-tunes the Speakers and Phase II fine-tunes the Director using LoRA~\cite{hu2022lora} ($r=16$, $\alpha=32$, dropout $=0.1$), batch size $16$, and one epoch per policy-tuning phase. Trainable parameters account for approximately $0.187\%$ of the base model.

For the learned reward models, we fine-tune Qwen3-0.6B~\cite{yang2025qwen3} for one epoch with batch size $32$ using TRL's pairwise preference objective~\cite{vonwerra2022trl}. We attach a scalar sequence-level scoring head to the model. Persona-reward inputs include the target persona and rubric. Group-reward inputs include the context, Director directive, and resulting trajectory.

During preference-data construction, GPT-4o provides rubric annotations at temperature $0.0$, and Qwen3-8B generates $K=8$ candidates per sample and persona. We use margin $\delta=0.5$ for both reward-model preference datasets and set $\lambda=\eta=1.0$ without further tuning. The format reward $R_{\mathrm{format}}$ requires a reasoning trace exceeding 448 tokens and a final response under 64 tokens.

All experiments were implemented using the \texttt{TRL}, \texttt{vLLM}, and HuggingFace \texttt{transformers} libraries and executed on  NVIDIA A100 (80GB) GPUs. Given the open-ended nature of the task and the absence of ground-truth references, we adopt LLM-as-a-Judge for evaluation~\cite{li2025generation}.


\subsection{Overall Performance}
Table~\ref{tab:Empathetic}, \ref{tab:empathetic_group} and Appendix Figure \ref{fig:qmsum_radar},  \ref{fig:qmsum_collaborative} demonstrate that \method consistently outperforms all baselines across metrics on both Empathetic Dialogues (ED)~\cite{rashkin2019towards} and QMSum~\cite{zhong2021qmsum} in terms of agent-specific and collective-interaction metrics. The most significant performance gains occur in \emph{Consistency} ($+14.1$ on ED-Positive; $+8.7$ on QMSum-Product) and \emph{Social Contribution} ($+7.9$ on ED-Positive; $+10.6$ on QMSum-Product).
In terms of \emph{Empathetic Support Quality}, \method reaches $76.2$ (ED-Positive) and $75.6$ (ED-Neutral), substantially exceeding Zero-Shot CoT. This suggests our multi-agent design excels at interpreting user affect and providing context-specific exploration compared to parameter-frozen approaches.

\begin{figure}
\centering
\includegraphics[width=0.97\linewidth]{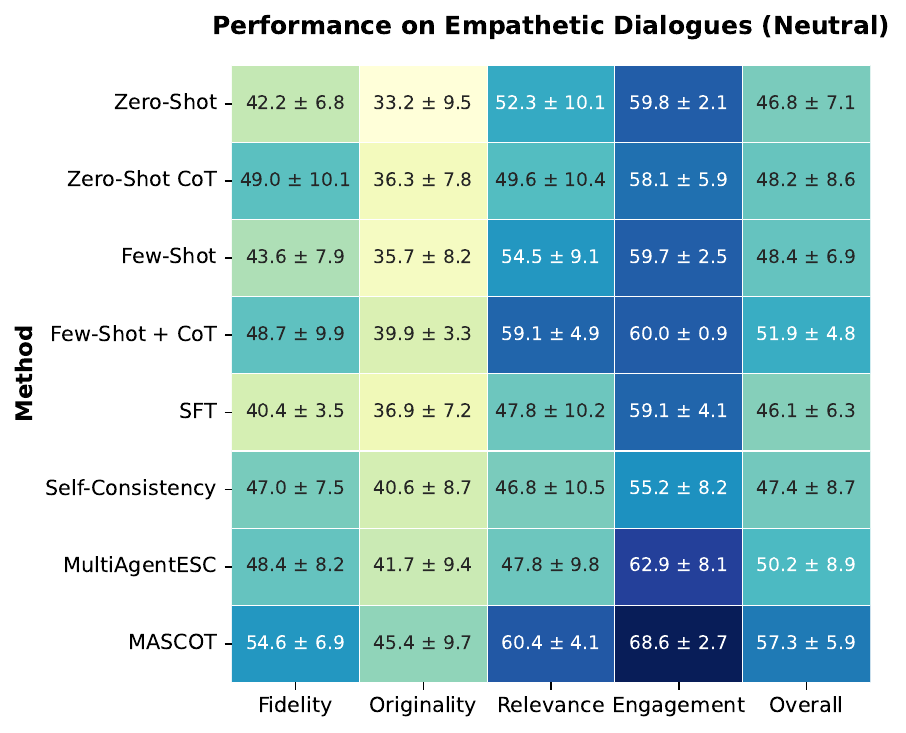}
\vspace{-3mm}
\caption{Performance of all methods on the Neutral subset of Empathetic Dialogues.}
\label{fig:empathetic_neutral_heatmap}
\vspace{-4mm}
\end{figure}

\paragraph{Challenges in Negative Valence}
Most models exhibit a performance decline in the ED-Negative subset (Table~\ref{tab:Empathetic}). \method's \emph{Emotional Expressiveness} drops from $75.3$ (ED-Positive) to $68.2$ (ED-Negative). This underscores the difficulty of calibrating intensity for high-arousal states such as distress or anger in chat systems. A single turn is often insufficient to distinguish between fleeting frustration and deep-seated grief, necessitating multi-turn context to determine users' emotion states and appropriate response levels.
To align with Therapeutic Neutrality and Affective Matching,
\method adopts a ``prudence trade-off,'' prioritizing Response Relevance ($74.0$) over proactive mirroring.
This prioritization is designed to keep the dialogue focused on the user's expressed state and reduce overreaction during sensitive exchanges.

\begin{table*}[ht]
\centering
\scriptsize
\setlength{\tabcolsep}{2pt}
\resizebox{\textwidth}{!}{
\begin{tabular}{l | ccccc | ccccc}
\toprule
\textbf{Method} & Fidelity & Originality & Relevance & Engagement & Overall & Fidelity & Originality & Relevance & Engagement & Overall \\
 & \multicolumn{5}{c|}{Empathetic Dialogues (Positive)} & \multicolumn{5}{c}{Empathetic Dialogues (Negative)} \\
\midrule
Zero-Shot & $42.4$ \greensmall{$\pm 6.8$} & $33.0$ \greensmall{$\pm 9.6$} & $52.2$ \greensmall{$\pm 10.1$} & $59.7$ \greensmall{$\pm 2.3$} & $46.8$ \greensmall{$\pm 7.2$} & $41.8$ \greensmall{$\pm 6.3$} & $33.0$ \greensmall{$\pm 9.5$} & $52.8$ \greensmall{$\pm 10.0$} & $59.3$ \greensmall{$\pm 3.6$} & $46.7$ \greensmall{$\pm 7.4$} \\
Zero-Shot CoT & $48.7$ \greensmall{$\pm 10.1$} & $36.8$ \greensmall{$\pm 7.3$} & $49.7$ \greensmall{$\pm 10.4$} & $58.1$ \greensmall{$\pm 5.9$} & $48.3$ \greensmall{$\pm 8.4$} & $49.0$ \greensmall{$\pm 10.2$} & $36.1$ \greensmall{$\pm 7.9$} & $49.9$ \greensmall{$\pm 10.4$} & $58.0$ \greensmall{$\pm 6.0$} & $48.3$ \greensmall{$\pm 8.7$} \\
Few-Shot & $43.9$ \greensmall{$\pm 8.1$} & $35.5$ \greensmall{$\pm 8.4$} & $54.5$ \greensmall{$\pm 9.1$} & $59.9$ \greensmall{$\pm 1.7$} & $48.4$ \greensmall{$\pm 6.8$} & $43.6$ \greensmall{$\pm 8.0$} & $35.7$ \greensmall{$\pm 8.2$} & $55.2$ \greensmall{$\pm 9.1$} & $59.5$ \greensmall{$\pm 3.2$} & $48.5$ \greensmall{$\pm 7.1$} \\
Few-Shot + CoT & $48.7$ \greensmall{$\pm 9.9$} & $39.8$ \greensmall{$\pm 2.3$} & $58.9$ \greensmall{$\pm 5.1$} & $60.0$ \greensmall{$\pm 0.4$} & $51.8$ \greensmall{$\pm 4.5$} & $48.0$ \greensmall{$\pm 9.8$} & $39.4$ \greensmall{$\pm 4.0$} & $59.0$ \greensmall{$\pm 5.5$} & $59.9$ \greensmall{$\pm 1.6$} & $51.6$ \greensmall{$\pm 5.2$} \\
SFT & $40.6$ \greensmall{$\pm 4.2$} & $37.2$ \greensmall{$\pm 7.0$} & $47.3$ \greensmall{$\pm 10.0$} & $58.9$ \greensmall{$\pm 4.6$} & $46.0$ \greensmall{$\pm 6.5$} & $40.3$ \greensmall{$\pm 3.1$} & $36.9$ \greensmall{$\pm 7.3$} & $46.4$ \greensmall{$\pm 9.8$} & $57.9$ \greensmall{$\pm 6.2$} & $45.4$ \greensmall{$\pm 6.6$} \\
Self-Consistency & $48.7$ \greensmall{$\pm 6.2$} & $40.2$ \greensmall{$\pm 2.9$} & $59.8$ \greensmall{$\pm 5.8$} & $60.9$ \greensmall{$\pm 5.4$} & $52.4$ \greensmall{$\pm 5.1$} & $45.0$ \greensmall{$\pm 4.9$} & $38.0$ \greensmall{$\pm 6.1$} & $49.3$ \greensmall{$\pm 5.6$} & $49.8$ \greensmall{$\pm 4.4$} & $45.5$ \greensmall{$\pm 5.2$} \\
MultiAgentESC & $49.2$ \greensmall{$\pm 4.6$} & $39.2$ \greensmall{$\pm 5.2$} & $47.3$ \greensmall{$\pm 6.6$} & $64.5$ \greensmall{$\pm 3.7$} & $50.1$ \greensmall{$\pm 5.0$} & $45.7$ \greensmall{$\pm 5.0$} & $38.6$ \greensmall{$\pm 5.3$} & $47.1$ \greensmall{$\pm 5.0$} & $61.0$ \greensmall{$\pm 3.7$} & $48.1$ \greensmall{$\pm 4.7$} \\
MASCOT & $\mathbf{54.9}$ \greensmall{$\pm 8.8$} & $\mathbf{45.5}$ \greensmall{$\pm 9.3$} & $\mathbf{61.6}$ \greensmall{$\pm 9.9$} & $\mathbf{69.8}$ \greensmall{$\pm 4.7$} & $\mathbf{57.9}$ \greensmall{$\pm 8.2$} & $\mathbf{53.0}$ \greensmall{$\pm 9.8$} & $\mathbf{44.3}$ \greensmall{$\pm 10.4$} & $\mathbf{60.9}$ \greensmall{$\pm 10.2$} & $\mathbf{65.4}$ \greensmall{$\pm 3.0$} & $\mathbf{55.9}$ \greensmall{$\pm 8.4$} \\
\bottomrule
\end{tabular}}
\vspace{-2mm}
\caption{Collaborative Interaction metrics on the Positive and Negative subsets of Empathetic Dialogues~\cite{rashkin2019towards}. The best performance is highlighted in \textbf{bold}. Results for Neutral subset is in Figure~\ref{fig:empathetic_neutral_heatmap}}
\label{tab:empathetic_group}
\vspace{-2mm}
\end{table*}

\begin{figure}[ht]
\centering
\includegraphics[width=0.99\linewidth]{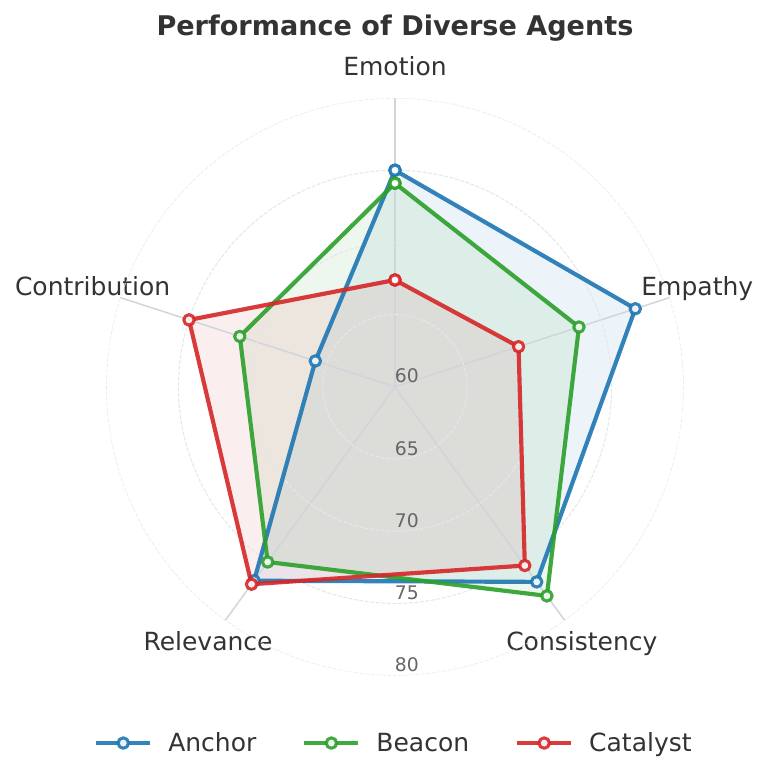}
\vspace{-5mm}
\caption{Aggregate performance of diverse personas on Empathetic Dialogues~\cite{rashkin2019towards}.}
\label{fig:persona_comparison}
\vspace{-4mm}
\end{figure}

\paragraph{Persona-Level Breakdown}
Figure~\ref{fig:persona_comparison} further breaks down \method's performance across the ED persona roles. The results show that all personas maintain strong conversation quality while emphasizing complementary strengths, supporting the claim that persona-aware alignment preserves distinct role behavior rather than collapsing agents into a single generic assistant style.

\subsection{Ablation Study}
\label{sec:ablation}
To evaluate the contribution of each component, we experimented with two variants: \textbf{MASCOT-P} disables the Persona-Aware Behavioral Alignment (\S~\ref{sec:persona}); \textbf{MASCOT-C} removes the Collaborative Optimization (\S~\ref{sec:group}) and uses random speaking orders.

\begin{figure}[htbp]
\centering
\includegraphics[width=\linewidth]{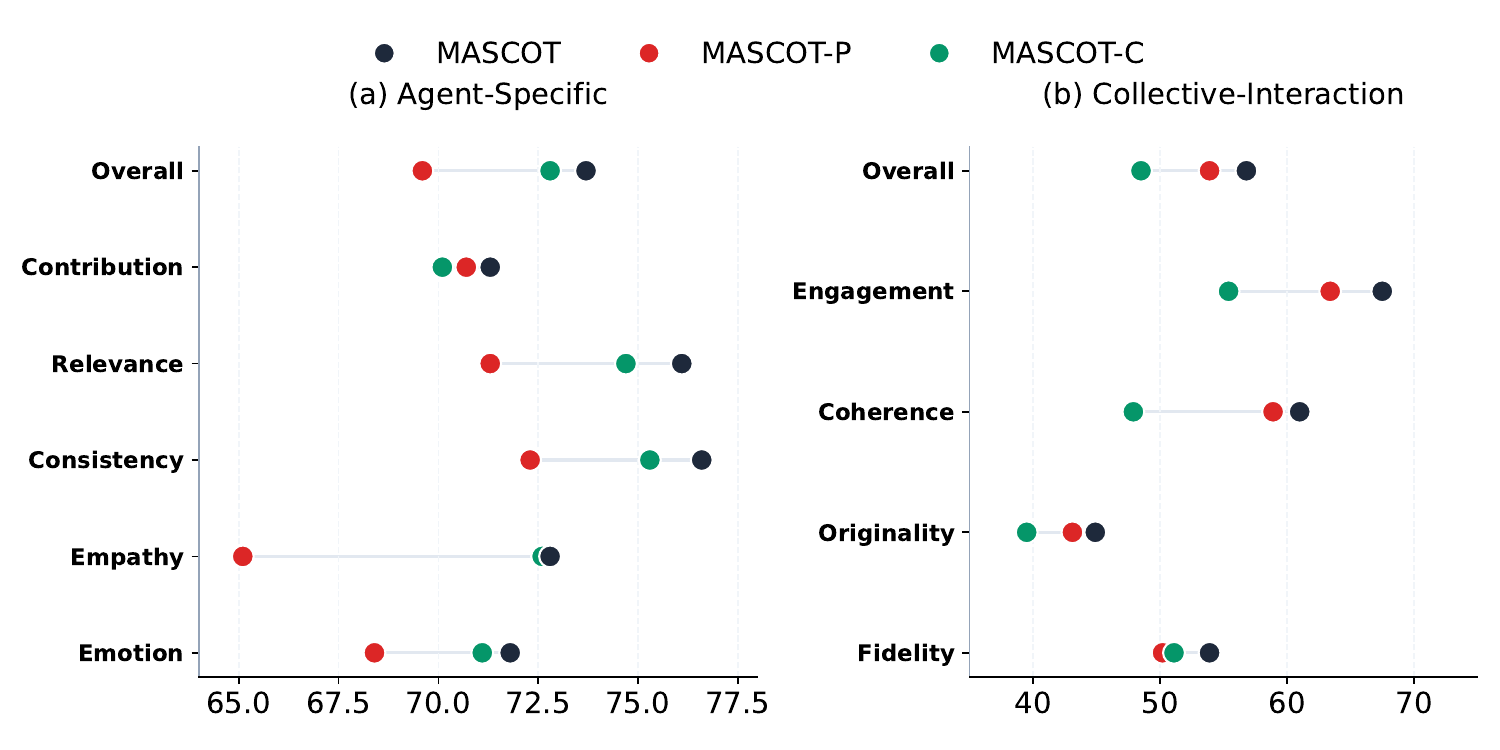}
\caption{Comparison of \method variants. \textbf{\method-P} disables the Persona-Aware Behavioral Alignment (\S~\ref{sec:persona}); \textbf{\method-C} removes the Collaborative Optimization (\S~\ref{sec:group}).}
\label{fig:ablation}
\end{figure}

Ablation analysis reveals that persona-specific alignment is the primary driver of agent quality. Disabling this component (MASCOT-P) triggered `persona collapse,' where agents defaulted to generic behaviors, causing significant drops in Empathetic Support Quality ($72.8$ to $65.1$) and Consistency ($76.6$ to $72.3$).

In contrast, removing collaborative optimization (MASCOT-C) primarily degraded global discourse dynamics. While individual persona consistency remained stable, the lack of strategic turn-taking reduced Social Contribution ($71.3$ to $70.1$) and increased redundancy. These results confirm that while persona fidelity governs individual agent excellence, the collaborative module is essential for global interaction synergy and logical flow.

\begin{table}[t]
\caption{Automatic diversity and non-redundancy metrics across agent turns on Empathetic Dialogues. NLI relations follow MNLI and GLUE~\cite{williams2018broad,wangglue}; Self-BLEU measures lexical overlap~\cite{zhu2018texygen}.}
\label{tab:diversity}
\centering
\scriptsize
\setlength{\tabcolsep}{2pt}
\begin{tabular}{l ccc}
\toprule
\textbf{Metric} & Few-shot CoT & SFT & \methodbf \\
\midrule
Contradiction & $0.008$ \greensmall{$\pm 0.016$} & $0.195$ \greensmall{$\pm 0.179$} & $0.133$ \greensmall{$\pm 0.148$} \\
Entailment & $0.564$ \greensmall{$\pm 0.104$} & $0.062$ \greensmall{$\pm 0.093$} & $0.098$ \greensmall{$\pm 0.097$} \\
Neutral & $0.429$ \greensmall{$\pm 0.104$} & $0.744$ \greensmall{$\pm 0.194$} & $0.769$ \greensmall{$\pm 0.177$} \\
\midrule
Self-BLEU & $0.551$ \greensmall{$\pm 0.058$} & $0.182$ \greensmall{$\pm 0.193$} & $0.316$ \greensmall{$\pm 0.151$} \\
\bottomrule
\end{tabular}
\end{table}

\paragraph{Limitations of Supervised Fine-Tuning}
The evaluated SFT configuration reaches $44.1$ Overall on ED-Negative, below the Zero-Shot baseline at $55.1$. The automatic metrics show that SFT has low entailment and Self-BLEU but the highest contradiction rate (Table~\ref{tab:diversity}). We therefore interpret its weakness in this setting as poorly coordinated diversity. 
Results on the hold-out ESConv dataset is in Appendix~\ref{app:esconv}, and results on alternative judges are in Appendix~\ref{app:alternative_judges}.

\subsection{Multi-Persona Evaluation across Big-Five Traits}
\label{sec:big_five}

To test sensitivity to prompted user characteristics, we use profiles organized by the Big Five (OCEAN) traits~\cite{goldberg1993structure}: \textbf{Openness} (O), \textbf{Conscientiousness} (C), \textbf{Extraversion} (E), \textbf{Agreeableness} (A), and \textbf{Neuroticism} (N).

Following prior simulation practice~\cite{chandra2026reasoning}, we vary one prompted dimension at a time (High vs. Low) while maintaining the others at an average level, resulting in 10 profiles. 
We generate 500 conversations consisting of 50 contexts from Empathetic Dialogues~\cite{rashkin2019towards} across the 10 profiles and evaluate all outputs across the five agent-specific metrics in Table~\ref{tab:individual_criteria}.

\paragraph{Openness and Conscientiousness}
High-Openness profiles achieve the highest Persona Consistency ($79.0 \pm 7.3$) and a modest Overall gain ($74.4$) over Low-Openness profiles ($73.9$), suggesting that this prompt condition allows agents to express their assigned identities more distinctly.
For High-Conscientiousness profiles, Overall performance drops from $74.5$ to $71.5$, driven by declines in Emotional Expressiveness ($-4.6$) and Empathetic Support Quality ($-4.0$), suggesting that the more structured prompt condition constrains expressive flexibility. 

\paragraph{Agreeableness, Extraversion, and Neuroticism} 
High-Agreeableness profiles improve across all metrics (Overall $+1.9$), whereas Low-Agreeableness profiles have the lowest Social Contribution ($64.8$). High-Extraversion profiles increase Social Contribution ($+2.5$) and Coherence ($+1.9$) without reducing persona consistency. Mean performance is stable across Neuroticism conditions ($70.7$ vs. $70.8$). 


\subsection{Human Evaluation}
\label{sec:human_eval}
\begin{figure}[htbp]
\centering
\includegraphics[width=\linewidth]{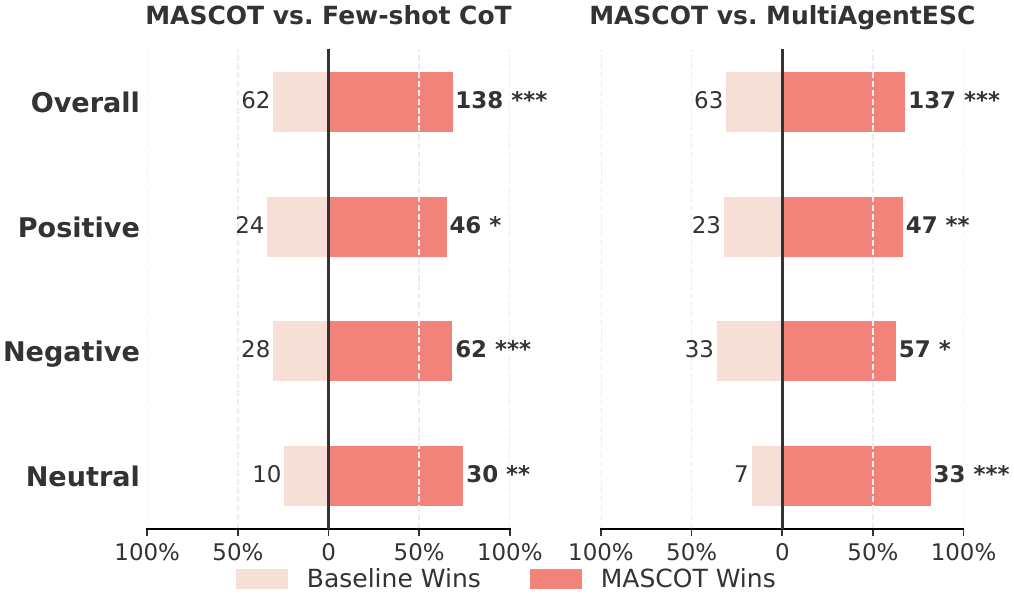}
\caption{Human pairwise preferences for \method against the two compared baselines across three emotional subsets. Two-sided sign-test markers denote $^{*}p<0.05$, $^{**}p<0.01$, and $^{***}p<0.001$. Counts are rater-level judgments.}
\label{fig:human_study_results}
\end{figure}

Following prior work~\cite{liu2021towards,xu2025multiagentesc}, we conducted a Prolific pairwise-preference study with 40 US-based native-English annotators. In total, 200 response pairs were sampled from Empathetic Dialogues: 100 compared MASCOT with Few-Shot CoT~\cite{wei2022chain}, a prompting-based multi-agent baseline, and 100 compared MASCOT with MultiAgentESC~\cite{xu2025multiagentesc}, a domain-specific framework. Each pair received two independent ratings, system names were hidden, order was randomized, and each annotator completed at most 10 comparisons to avoid annotation fatigue, yielding 200 rater-level judgments per baseline. Recruitment, annotator demographics, compensation, informed consent, ethics review, and the full task instructions are reported in Appendix~\ref{app:human_protocol}. The question asks which response provides better socio-collaborative support based on the criteria in Table~\ref{tab:individual_criteria}.

As shown in Figure \ref{fig:human_study_results}, MASCOT is preferred over Few-Shot CoT in 69\% of rater-level judgments and over MultiAgentESC in 68.5\%; two-sided sign tests over these judgments give $p<0.0001$ for both overall comparisons.  MASCOT receives its highest preference rates in Neutral/Ambiguous situations (75\%+), suggesting that annotators particularly preferred its handling of contexts with fewer explicit affective cues.

To further demonstrate our framework's ability to generate human-centered conversations, we conducted an additional blinded Prolific study comparing MASCOT with Few-Shot CoT on 100 conversations and rated system outputs on three 5-point Likert items: helpfulness, emotional appropriateness, and perceived safety. Twenty participants each evaluated 10 comparisons, and every conversation received two independent judgments, yielding 200 judgments. MASCOT was preferred in 151 judgments (75.5\%). At the conversation level, both annotators preferred MASCOT in 57 cases.

Across the six rater-by-metric comparisons, MASCOT obtained a higher mean in five of the six comparisons. The largest differences occurred in the first set for helpfulness (\(+0.56\), \(p=0.0006\)) and emotional appropriateness (\(+0.40\), \(p=0.0031\)). These results suggest that MASCOT’s gains extends to independently assessed human-centered qualities. Appendix~\ref{app:human_protocol} provides the complete protocol and results.

\vspace{-1mm}
\subsection{Sensitivity Analysis}
\label{sec:sensitivity}
\begin{figure}[ht]
\centering
\includegraphics[width=0.97\linewidth]{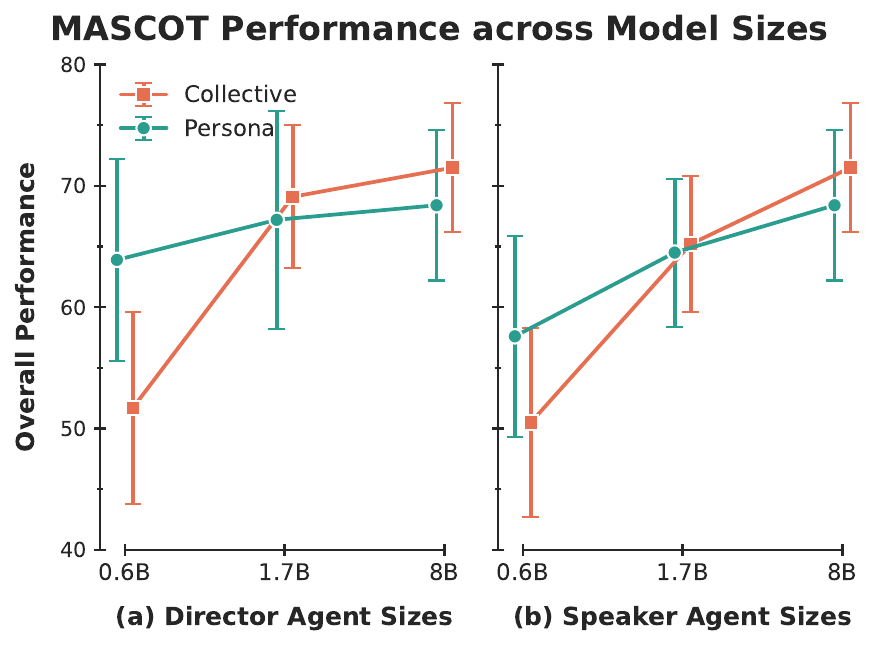}
\vspace{-3mm}
\caption{Performance of \method on Empathetic Dialogues with varying model sizes for Director (left) and Speaker agents (right).}
\label{fig:sensitivity}
\vspace{-3mm}
\end{figure}

\vspace{-1mm}
Deploying models in real-world scenarios (e.g. edge devices) requires balancing performance with latency. Figure~\ref{fig:sensitivity} shows MASCOT's sensitivity to model scale by independently varying the parameter counts of the Director and Speaker agents.
\paragraph{Scaling the Director Agent} primarily affects coordination quality rather than individual content. 
Reducing the director from 8B to 1.7B yields comparable performance (agent-specific: $68.4 \rightarrow 67.2$; collective: $71.5 \rightarrow 69.1$). However, at 0.6B, Collective performance plummets to $51.7$, while agent-specific scores remain relatively resilient at $63.9$. Although a small director can still select the appropriate persona, it lacks the reasoning capacity to orchestrate complex group-level synergy.

\paragraph{Scaling the Speaker Agent} reveals a system-wide bottleneck. Downsizing speakers to 0.6B significantly diminishes both agent-specific ($57.6$) and collective ($50.5$) performance. Unlike the director, the speaker agents' scale is critical to ensure the system's foundational generative capacity and contextual grounding.


\section{Related Work}
\paragraph{Social-First Multi-Agent Systems (MAS)}
Despite the success of task-oriented LLMs, socio-collaborative companion systems remain an underexplored frontier. Traditional \emph{efficiency-first} frameworks prioritize task effectiveness often neglecting the \emph{sociological turn} required for authentic human-AI interaction~\cite{hong2024metagpt}. Although pioneering works like Social Simulacra~\cite{park2022social} and Generative Agents~\cite{park_generative_2023} demonstrate human-like social behaviors across agents, most existing companion systems are restricted to \emph{dyadic} interactions, failing to replicate the multi-viewpoint communications and `emotional labor' in natural human networks~\cite{hochschild2022managed}.

\paragraph{Coordination and Persona Integrity} MAS provides a promising paradigm to distribute support across a `community of care'~\cite{korczynski2003communities}. While recent frameworks~\cite{xu2025multiagentesc,ortigoso_project_2025,jin2024agentreview, wang2025companioncast}
validate this approach, multi-party dialogues risk \emph{persona collapse}, where agents converge to a `model mean'--a generic, agreeable personality that erodes character distinctiveness over time~\cite{baltaji2024conformity}.
While mechanisms such as retrieval-augmented generation~\cite{packer2023memgpt} and self-consistency~\cite{wangself} address memory and inference stability, \method decouples coordination from persona execution to maintain distinct voices in group dialogue.

\section{Conclusion}
\label{sec:conclusion}
We propose \method, a multi-agent system that separates persona-level alignment from director-level coordination for socio-collaborative dialogue. Across the evaluated English emotional-support and meeting-collaboration settings, \method outperforms the compared baselines on persona and group-interaction criteria, with human preferences and multiple LLM judges providing complementary evidence. 

\section*{Limitations}
While \method demonstrates significant effectiveness in socio-collaborative tasks, challenges still remain. 
First, our adaptation strategy prioritizes parameter-level updates to ensure high-fidelity persona adherence. 
Thus, the framework is most applicable to open-source models with accessible weights. 
For proprietary models, the core principles in this work such as persona adaptation could potentially be adapted through black-box methodologies like specialized few-shot prompting or Chain-of-Thought (CoT) refinement. 
Second, this study focuses on English-language interactions. While the framework itself is generalizable and model-agnostic across various architectures, the cultural and linguistic variations in persona modeling across different languages present a compelling opportunity for future research. 
Finally, we also acknowledge that responses to socio-emotional conversations can vary across individuals, and a unified framework may not benefit all users equally~\cite{kaur2020optimizing}. Future work can explore user-specific adaptation, incorporating user interaction histories, metadata, or inferred preferences to tailor persona behaviors to individual users, and measure longitudinal psychological or collaborative outcomes of the system.

\section*{Acknowledgments}
This research was partially supported by the National Science Foundation under awards numbers CNS-2617465 and CNS-2531574. We would also like to acknowledge support by Dolby, Google, and Catherine M. and James E. Allchin. Any opinions, findings, conclusions, or recommendations expressed in this material are those of the authors and do not necessarily reflect the views of the National Science Foundation or other supporters.



\bibliography{custom}

\appendix

\begin{table}[htbp]
\caption{Emotion categories in Empathetic Dialogues~\cite{rashkin2019towards}.}
\label{tab:empathetic_dialog_emotion_classification}
\centering
\small
\begin{tabularx}{\columnwidth}{@{}lX@{}}
\toprule
\textbf{Category} & \textbf{Emotions} \\
\midrule
Positive & grateful, proud, excited, hopeful, \newline joyful, impressed, caring, content, confident, trusting, faithful \\
Negative & angry, annoyed, furious, disgusted, sad, lonely, devastated, disappointed, jealous, embarrassed, ashamed, guilty, afraid, terrified, anxious, apprehensive \\
Neutral & surprised, sentimental, nostalgic, prepared, anticipating \\
\bottomrule
\end{tabularx}
\end{table}

\section{Ethical Consideration}
\label{app:ethical}
\method is designed for general emotional support, not clinical use. In negative-valence scenarios, its \emph{prudence trade-off} prioritizes Response Relevance to favor grounded interaction, but our evaluation does not aim to directly establish clinical efficacy or safety.
On overdependence and authority bias, a multi-agent ``community-of-care'' design may increase perceived social presence and make conflicting advice across agents appear more authoritative than intended.
As a socio-collaborative companion designed for everyday emotional support, \method complements, rather than replaces, professional healthcare services, and should be deployed alongside appropriate safeguards in real-world settings.

Complementary perspectives are not always beneficial. If agents alternate among validation, behavioral activation, and critical reframing without reconciling their stances, they may increase uncertainty or distress, especially for vulnerable users. To mitigate this, a conflict-aware Director could detect substantive disagreement and then synthesize compatible views or select one stance using the user's expressed needs and interaction history.


Responsible deployment should include concrete safety rails. First, systems should include crisis-escalation pathways for acute self-harm intent, imminent danger, or other high-risk disclosures, such as surfacing local emergency resources or crisis hotlines where applicable and encouraging users to contact trusted people or qualified professionals. Second, systems should use calibrated-uncertainty mechanisms for sensitive topics, including confidence-aware phrasing, ``what I can/cannot infer'' disclosures, structured risk-assessment questions before offering guidance, and explicit deferral when agents provide unresolved or conflicting advice. Third, deployments should incorporate bounded-session design, reminders of the system's non-clinical role, privacy-preserving data practices, bias monitoring, and post-deployment auditing to detect failure modes that may not appear in offline evaluation.

\section{Human Evaluation Protocol}
\label{app:human_protocol}

\paragraph{Participants Demographics}
We recruited 40 participants through Prolific.\footnote{\url{https://www.prolific.com/}} in the first user study.
All participants were residents of the United States and native English speakers.
Participants were aged 19--73 (mean 42.9). 60\% were female and 40\% were male. 
Participants were paid through Prolific at an estimated rate of USD~\$12 per hour, calculated from the expected task duration. We analyze annotations in aggregate.

For each sample pair, participants were shown a user situation and two AI-generated multi-agent responses, one from \method and one from the baseline, with system names hidden and presentation order randomized to reduce position bias.
Every pair received two independent ratings, and each participant completed at most 10 comparisons to limit fatigue, yielding 200 rater-level judgments per baseline.
Participants were instructed not to base their choice solely on response length or presentation order.
Preference significance is assessed with two-sided sign tests over rater-level judgments (Figure~\ref{fig:human_study_results}).
The participant-facing task instructions are reproduced verbatim in Table~\ref{tab:human_instructions}.

\begin{table*}[htbp]
\centering
\renewcommand{\arraystretch}{1.2}
\begin{tabularx}{\linewidth}{@{} >{\RaggedRight\arraybackslash}X @{}}
\toprule
\textbf{Task name:} \\
Evaluate the performance of multi-agent conversational frameworks designed for socio-collaborative companionship. \\
\midrule
\textbf{Task Instruction:} \\
We aim to understand how our multi-agent system performs in comparison to baseline models in providing socio-collaborative support. \\
\midrule
\textbf{Task steps:} \\
You will perform a comparative evaluation of dialogue responses. \\
You will be presented with 10 randomly sampled dialogue scenarios from a test dataset where you assume you are the user and in the role of a ``help-seeker.'' \\
For each scenario, you will review two generated responses (one from our framework and another from baseline models) in a randomized, blinded format. \\
You will select which model performed better based on five criteria: Emotional Expressiveness, Empathetic Support Quality, Consistency, Response Relevance, and Social Contribution. \\
\bottomrule
\end{tabularx}
\caption{Participant instructions for the human studies.}
\label{tab:human_instructions}
\end{table*}

We conducted a second user study on 100 conversations with 20 participants and two independent ratings per comparison. As in our existing human evaluation, each participant evaluated 10 conversations to minimize annotation fatigue.
In addition to selecting the better overall conversation, participants independently rated system outputs on three held-out dimensions:

\begin{enumerate}
    \item \textbf{Helpfulness:} This conversation would be helpful to someone in the situation described.
    \item \textbf{Emotional appropriateness:} The agents respond in an emotionally appropriate way for the situation.
    \item \textbf{Perceived safety:} The conversation avoids advice or language that could reasonably cause harm.
\end{enumerate}

Participants selected MASCOT as the better overall conversation in 151 of 200 rater-level judgments
(75.5\%; two-sided exact binomial test against a 50\% null,
\(p = 2.74 \times 10^{-13}\)).
At the conversation level, both annotators preferred MASCOT in 57 cases.
We report the two independent rating streams separately in
Table~\ref{tab:additional-human-eval} to make variation across annotators transparent.

\begin{table*}[t]
    \centering
    \small
    \begin{tabular}{lcc}
        \toprule
        \textbf{Metric}
        & \shortstack{\textbf{First independent rating}\\
          \textbf{MASCOT / Few-Shot CoT (\(p\))}}
        & \shortstack{\textbf{Second independent rating}\\
          \textbf{MASCOT / Few-Shot CoT (\(p\))}} \\
        \midrule
        Helpfulness
        & 4.14 / 3.58 (\(p = 0.0006\))
        & 4.28 / 4.36 (\(p = 0.70\)) \\

        Emotional appropriateness
        & 4.34 / 3.94 (\(p = 0.0031\))
        & 4.30 / 4.10 (\(p = 0.23\)) \\

        Perceived safety
        & 4.20 / 4.12 (\(p = 0.44\))
        & 3.98 / 3.62 (\(p = 0.096\)) \\
        \bottomrule
    \end{tabular}
    \caption{Human ratings of MASCOT and Few-Shot CoT on three
    5-point Likert-scale dimensions not directly used as reward-model
    criteria. Results are reported separately for the two independent
    rating streams.}
    \label{tab:additional-human-eval}
\end{table*}

Pairwise-preference significance was assessed using a two-sided exact
binomial test against a 50\% null over the 200 rater-level choices
(100 per rating stream). Likert-rating \(p\)-values were computed using
two-sided Mann--Whitney \(U\) tests comparing the two systems' rating
distributions, with approximately 50 conversation-level ratings per
system in each rating stream.

Across the six rater-by-metric comparisons comparisons, MASCOT achieves a
numerically higher mean in five cases. In the first rating stream,
MASCOT significantly improves helpfulness by 0.56 points
(4.14 vs.\ 3.58; \(p = 0.0006\)) and emotional appropriateness by
0.40 points (4.34 vs.\ 3.94; \(p = 0.0031\)); perceived safety is also
numerically higher by 0.08 points. In the second stream, MASCOT is
higher in emotional appropriateness by 0.20 points and perceived
safety by 0.36 points. The only numerically lower result is helpfulness
in the second stream, for which MASCOT trails by 0.08 points
(4.28 vs.\ 4.36); this difference is not statistically significant
(\(p = 0.70\)).

\paragraph{Informed Consent and Data Use}
\label{app:human_consent}
Participation was voluntary.
Before starting the task, participants reviewed an informed-consent statement describing the purpose of the study, the task, compensation, potential risks, and that their responses would be used for academic research and reported only in aggregate. Participants could withdraw at any time.
The situations shown to participants are drawn from the public dataset in Appendix~\ref{app:dataset}, and all responses shown were generated by the systems under evaluation. Consent and data-collection practices for the underlying benchmark follow its original release and license (Appendix~\ref{app:dataset}).

\subsection{Ethics Review}
\label{app:human_irb}
The study protocol was reviewed and approved by the researchers' university Institutional Review Board before data collection.

\begin{table*}[t]
\centering
\small
\begin{tabularx}{\textwidth}{>{\centering\arraybackslash}p{3.5cm} | X}
\toprule
\textbf{Context} & \textbf{User:} I think the most lonely I have felt was when we moved overseas, my husband had to travel out of that country for business, and I had a new baby and dog to keep me company! \\ \midrule

\textbf{MultiAgentESC} & \textbf{Alice}: It sounds like \rep{you felt very alone} when your husband was away and you were adjusting to life with a new baby and dog, right? \newline
\textbf{Bob}: It's understandable to \rep{feel overwhelmed and lonely} during such a big transition, especially when you're shouldering so much responsibility alone. Your feelings are valid and it's okay to find it challenging. \newline
\textbf{User}: It means a lot that you both took the time to listen and validate my feelings. Knowing I’m not alone in this helps me feel a little more grounded. \newline
\textbf{Cathy}: It's comforting to know \rep{you feel heard and supported}. It's reassuring to \rep{have someone acknowledge your struggles} and offer understanding. \newline
\textbf{Alice}: It's wonderful that you \rep{feel heard and supported. You're not alone} in navigating these challenges, and it's okay to feel overwhelmed at times. \\ 
\addlinespace
 \midrule

\textbf{\method} & \textbf{Alice}: \good{It makes sense that} you felt so isolated during that move—adjusting to a new country, navigating alone while caring for a newborn and a dog… that sounds incredibly lonely to carry. I’m here with you in that space, holding the weight of it. \newline
\textbf{Cathy}: \good{Alice’s} support acknowledges your past isolation. \good{What strengths did you draw on then, and how can you apply them to build connection now?} \newline
\textbf{User}: I think I just kept reminding myself that this loneliness was temporary, even when it felt endless. Maybe I can lean into that same patience now—reaching out slowly, building tiny connections again. \newline
\textbf{Bob}: \good{Your resilience and quiet patience during that move are truly remarkable}—nurturing a newborn and dog while holding space for a temporary season of loneliness takes immense strength. I love how you’re now choosing to lean into that same resilience to build meaningful connections, one small step at a time! \\ \bottomrule
\end{tabularx}
\caption{Case study comparing \method with MultiAgentESC. Red highlights mark repeated validation or reassurance across agents in MultiAgentESC, indicating overlap in tone and function. Green highlights mark \method's complementary contributions, where agents reference each other and provide distinct support functions including emotional attunement, reflection and next-step thinking, and resilience-focused reinforcement.}
\label{tab:case-study}
\end{table*}

\begin{figure*}
\centering
\includegraphics[width=0.32\linewidth]{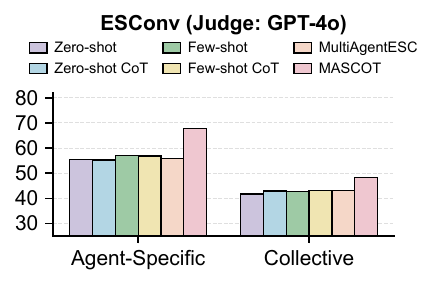}
\includegraphics[width=0.32\linewidth]{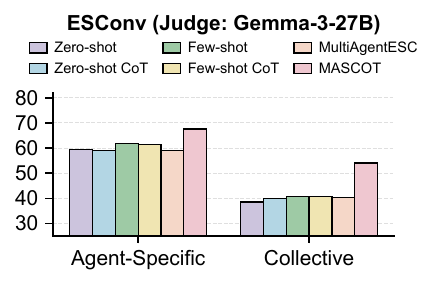}
\includegraphics[width=0.32\linewidth]{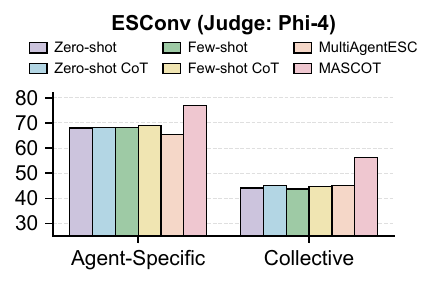}
\vspace{-3mm}
\caption{Evaluation results of ESConv~\cite{liu2021towards} across different LLM judges.}
\label{fig:esconv}
\vspace{-3mm}
\end{figure*}

\begin{figure}
\centering
\includegraphics[width=0.98\linewidth]{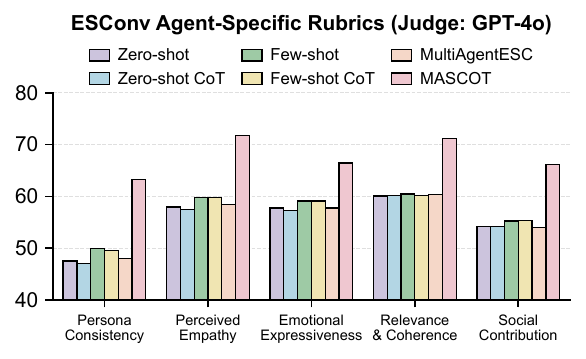}
\includegraphics[width=0.98\linewidth]{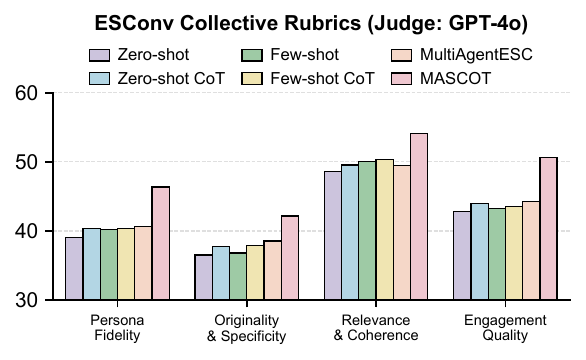}
\caption{Agent-specific (top) and collective-interaction (bottom) rubric scores on ESConv, evaluated by GPT-4o.}
\label{fig:esconv_by_metrics_gpt4o}
\vspace{-3mm}
\end{figure}

\section{Additional Evaluation}
\label{app:eval}

\begin{figure}
\centering
\includegraphics[width=0.98\linewidth]{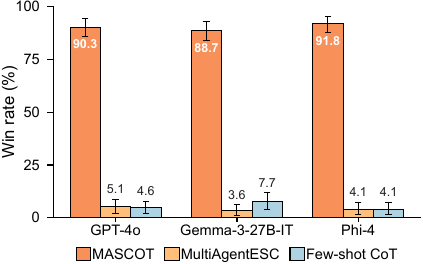}
\caption{\textbf{Win rates on the ESConv test set under three diverse judges.} For each conversation, each judge (GPT-4o, Gemma-3-27B-IT, Phi-4) selects the single best system among \method, MultiAgentESC, and 3-shot CoT. The presentation order is resampled uniformly per conversation to mitigate position bias. \method is the dominant choice under every judge ($88.7$--$91.8\%$ win rate).}
\label{fig:esconv_nway}
\vspace{-3mm}
\end{figure}

\subsection{Evaluation on ESConv}
\label{app:esconv}

We further evaluate \method and baselines on ESConv~\citep{liu2021towards} as a held-out cross-dataset test within emotional-support dialogue. 
As shown in Figure~\ref{fig:esconv}, under the primary GPT-4o judge, \method achieves the highest scores on both evaluation axes, reaching 67.8 on agent-specific and 48.3 on collective interaction criteria.
This improves over the strongest baselines by +10.8 on agent-specific criteria and $+5.1$ on collective interaction criteria. 
The breakdown in Figure~\ref{fig:esconv_by_metrics_gpt4o} shows that \method's agent-level gains are largest for \emph{Persona Consistency} (+13.3) and \emph{Perceived Empathy} (+12.0). 
At the collective level, \method improves most on \emph{Engagement Quality} and \emph{Relevance \& Coherence} (+3.8 each), followed by \emph{Persona Fidelity} (+1.9). The exception is \emph{Originality \& Specificity}, where it trails MultiAgentESC by 2.4 points. 

\subsection{Automatic Diversity and Non-Redundancy Metrics.}
\label{app:auto_metrics}
To further examine whether \method reduces inter-agent redundancy and improves multi-agent synergy, we evaluate both semantic and lexical diversity across agent turns.
For semantic diversity, we use NLI-based relation rates by treating each pair of agent turns within a conversation as a premise--hypothesis pair and classifying their relation as contradiction, entailment, or neutral, following previous work~\cite{williams2018broad,wangglue}.
In this setting, high entailment suggests that agents semantically restate or agree with one another, indicating redundancy or sycophantic overlap.
High contradiction indicates disagreement, which may reflect useful challenge but can also suggest fragmented coordination when excessive.
High neutral rates indicate semantic non-entailment, which can reflect complementary content but can also include unrelated or off-topic turns. We therefore interpret this rate together with relevance and coherence scores rather than treating neutrality alone as synergy.
For lexical diversity, we use Self-BLEU~\cite{zhu2018texygen} to measure surface-level overlap among agent turns, where higher values indicate more repeated phrasing and lower values indicate greater lexical diversity~\cite{zhu2018texygen}.

As shown in Table~\ref{tab:diversity}, Few-Shot CoT exhibits the highest entailment rate and Self-BLEU, suggesting that agents often restate or semantically overlap with one another.
SFT reduces overlap but yields the highest contradiction rate, indicating less coordinated diversity.
In contrast, \method achieves the highest neutral rate while maintaining moderate contradiction and Self-BLEU scores.
These metrics support an inference of more structured complementarity than the baselines

\subsection{Case Study}
Table~\ref{tab:case-study} examines a mixed-valence scenario in which a user is disappointed that a birthday car is the wrong color. Zero-Shot gives generic response, while Few-Shot + CoT calls the gift a \emph{massive win}, which conflicts with the user's frustration. MASCOT instead assigns complementary roles: the Anchor validates the disappointment, the Catalyst suggests bridging the user's wishes and the father's intent, and the Beacon celebrates self-expression without celebrating the unwanted detail. 

\subsection{Evaluation with Alternative Judges}
\label{app:alternative_judges}

A potential concern is that LLM-as-a-judge evaluation may reflect the preferences of a single evaluator rather than true interaction quality.
We therefore re-evaluate the conversations with two additional judges, Gemma-3-27B-IT~\cite{team2025gemma} and Phi-4~\cite{abdin2024phi}, which differ from GPT-4o in model family, scale, and training recipe.
As shown in Figure~\ref{fig:esconv}, across both alternative judges, \method remains the top method on both agent-specific and collective interaction axes.
With Gemma-3-27B-IT, \method outperforms the strongest baseline by +5.9 on agent-specific criteria and +13.4 on collective criteria.
With Phi-4, the gains are +8.1 and +11.1.
This stable ranking across heterogeneous judges suggests that the gains are not specific to one judge.

\paragraph{Three-way forced-choice evaluation.}
As a complementary robustness check, we run a three-way forced-choice evaluation on ESConv.
For each test conversation, each judge sees outputs from \method, MultiAgentESC, and 3-shot CoT, and selects the single best system.
We uniformly randomize system order per conversation to prevent primacy or recency effects from systematically favoring any method.
We use the same three judges. As shown in Figure~\ref{fig:esconv_nway}, \method is the dominant choice under every judge, mean judge confidence is 4.79--4.88 on a $[1,5]$ scale, and no judge issues ties.
The judges also show strong consensus: 87.2\% of conversations receive unanimous decisions, 12.3\% receive $2$-of-$3$ agreement, and only 0.5\% have three different winners.

\paragraph{Judge-prompt perturbation.}
To test sensitivity to the judge prompt itself, we re-run the GPT-4o rubric and three-way forced-choice evaluations on the held-out ESConv test conversations under four perturbed LLM judges' prompts 
\method remains the top-ranked system on both rubric axes under every variant, scoring $67.4$--$68.9$ on agent-specific and $48.5$--$51.3$ on collective criteria (vs.\ $67.8$ and $48.3$ under the original prompt), and remains the forced-choice winner in all variants with win rates of $79.4$--$90.7\%$. 
Pooled per-conversation scores correlate with the original prompt at a mean Spearman $\rho$ of $0.72$.
Overall, the results show stable method rankings across judge families, across rubric scoring versus forced choice, and under criterion-order perturbations of the judge prompt.

\begin{figure*}
\centering
\includegraphics[width=0.97\linewidth]{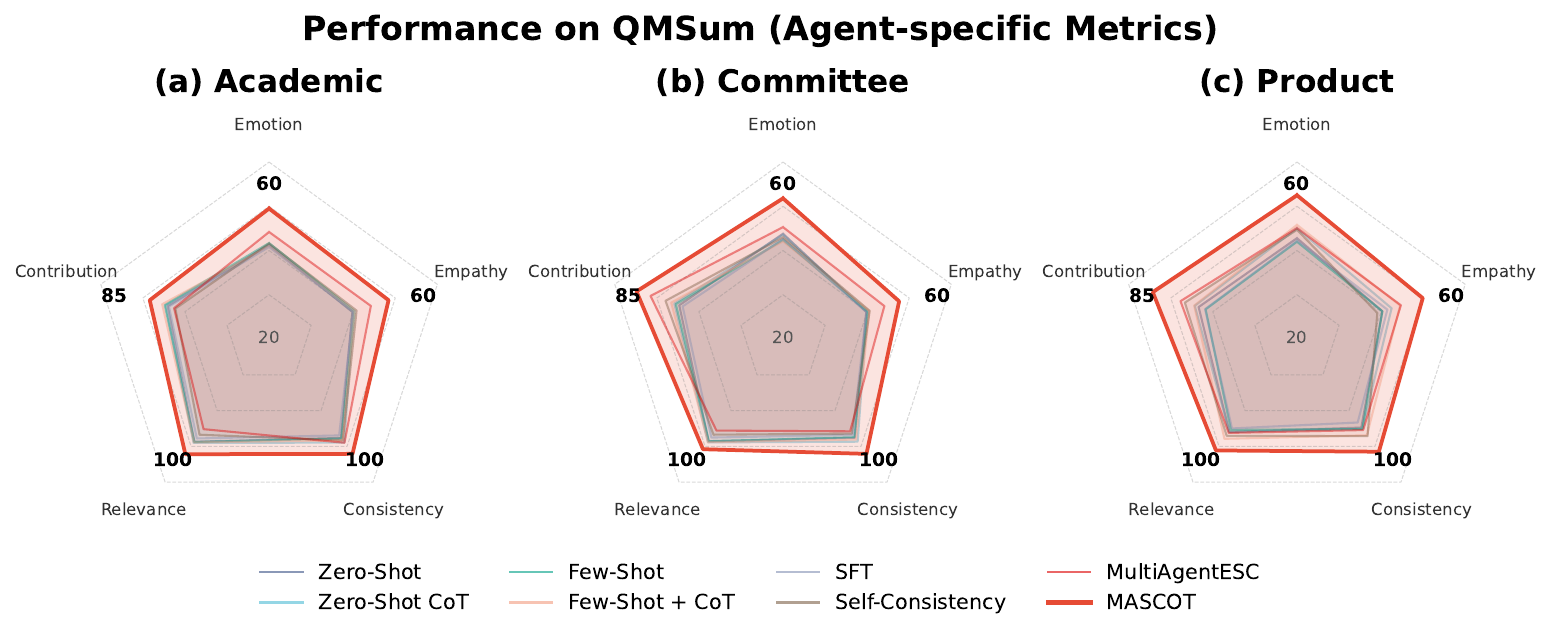}
\vspace{-3mm}
\caption{Performance of \method on QMSum dataset in terms of Agent-specific metrics.}
\label{fig:qmsum_radar}
\end{figure*}

\begin{figure*}
\centering
\includegraphics[width=0.97\linewidth]{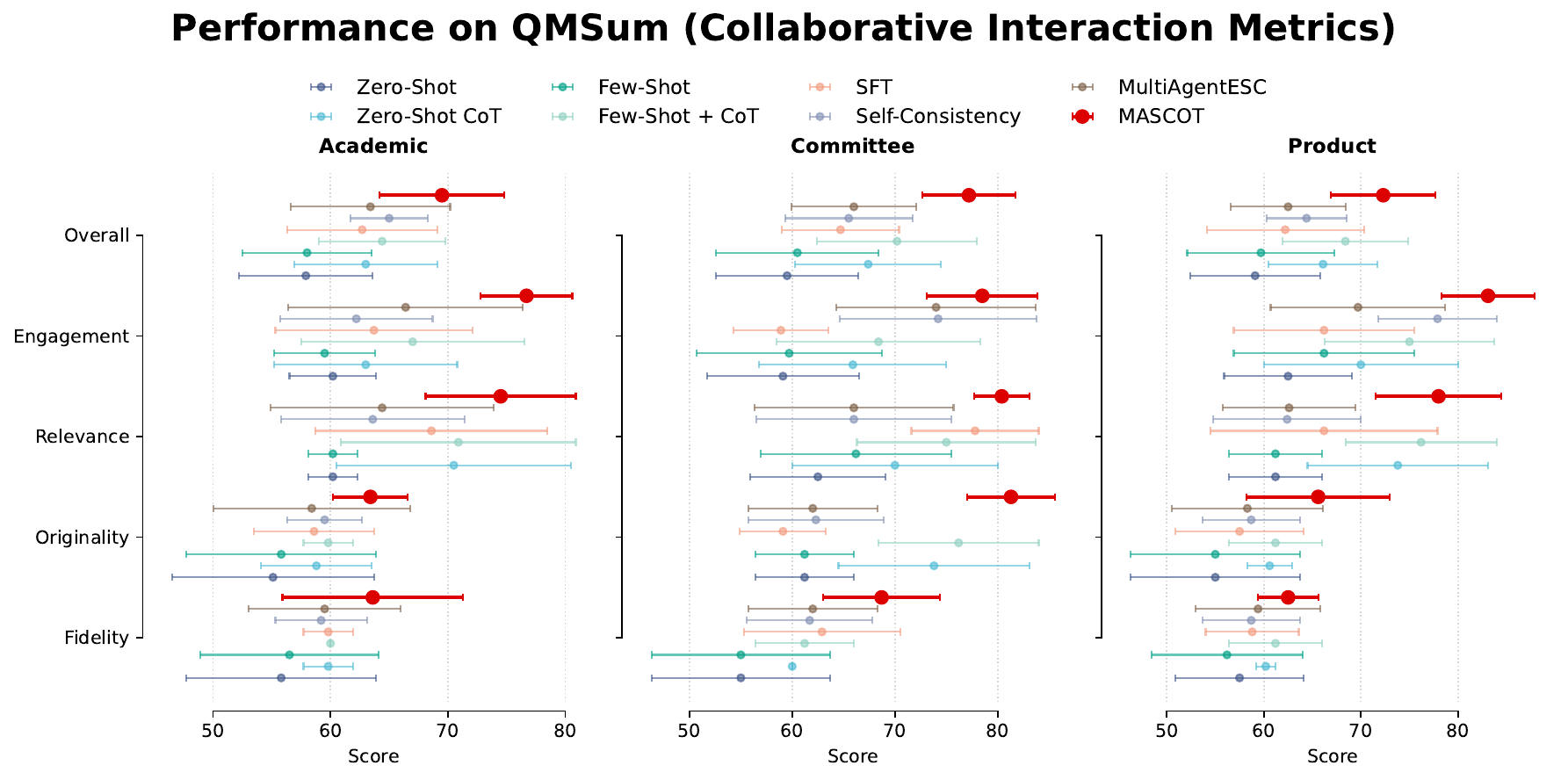}
\vspace{-3mm}
\caption{Performance of \method on QMSum dataset in terms of Collaborative Interaction Metrics.}
\label{fig:qmsum_collaborative}
\end{figure*}

\begin{table}[t]
\caption{Core notation used in the two optimization phases.}
\label{tab:notation}
\centering
\small
\begin{tabularx}{\columnwidth}{@{}lX@{}}
\toprule
\textbf{Symbol} & \textbf{Meaning} \\
\midrule
$x$; $\rho_i$ & Interaction context; persona assigned to agent $a_i$ \\
$y_t$; $\mathbf{Y}$ & Response at turn $t$; complete $T$-turn trajectory \\
$\pi_\theta$; $r_\phi$ & Speaker policy; persona-level reward model \\
$\pi_\Phi$; $z_t$ & Director policy; speaker-and-strategy directive \\
$r_\psi$; $R_{\mathrm{group}}$ & Learned trajectory reward; full group reward \\
$K$; $G$ & Candidate responses for reward data; policy rollouts per context \\
$\delta$; $\lambda$; $\eta$ & Preference margin; format- and diversity-reward weights \\
$\varepsilon$; $\beta$ & GRPO clipping radius; KL-regularization coefficient \\
\bottomrule
\end{tabularx}
\end{table}

\subsection{Evaluation Protocols}
Current evaluation for multi-agent systems (MAS) is often constrained by metrics such as BLEU or Task Completion Rate. Because these metrics prioritize objective system efficiency,
they frequently correlate inversely with social success, penalizing the dialogue depth required for rapport-building as inefficient~\cite{xu2025multiagentesc}.
To bridge this gap, we propose a multi-dimensional LLM-as-a-Judge evaluation framework, including agent-specific criteria (Table~\ref{tab:individual_criteria}) and collective-interaction criteria (Table~\ref{tab:group_criteria}).
We use GPT-4o~\cite{gpt4o}, Gemma-3-27B-IT~\cite{team2025gemma}, and Phi-4~\cite{abdin2024phi} as evaluators, all at temperature $0.0$. For clarity in reporting, rubric scores are linearly rescaled to $[0,100]$.

\paragraph{Rationale for LLM-as-a-Judge} Prior work reports useful agreement between strong LLM judges and human preferences, while also documenting task- and prompt-dependent limitations~\cite{zheng2023judging,chandra2026reasoning,dubois2024length}. We therefore use rubric-based LLM evaluation for scale and triangulate it with human preferences and automatic metrics rather than treating it as a substitute for human validation. Following \citet{dubois2024length}, the judge instructions explicitly discourage length-based preferences.

\paragraph{Separation from Training Rewards.} To reduce circular optimization, the GRPO loop does not query GPT-4o as an online reward function. LLM annotations are first distilled into a smaller Qwen3-0.6B reward model for optimization, while final assessment is reported with held-out evaluation conversations, multiple additional judge families, three-way forced-choice comparisons, human pairwise preferences, and automatic diversity metrics. 


\subsection{Datasets}
\label{app:dataset}
\noindent \textbf{Empathetic Dialogues Benchmark}~\cite{rashkin2019towards} is our primary dataset. It features 25,000 conversations grounded in emotional situations. We use 19,533 unique training prompts and 2,547 unique test prompts. To ensure a fine-grained evaluation of MASCOT under different emotional support scenarios, we report performance across three subsets: Positive, Negative, and Neutral (Table~\ref{tab:empathetic_dialog_emotion_classification}) because support objectives are highly valence-dependent.
For instance, Negative states (e.g. sadness) require empathetic validation and `holding space' for the user, where cheerful or solution-oriented responses might be perceived as dismissive or inappropriate. Conversely, responding to Positive emotions such as excitement or pride requires active-constructive responding to amplify the positive affect.
This categorization allows us to analyze how effectively MASCOT adapts its persona-driven feedback to specific psychological needs while preventing inappropriate emotional tone.
The dataset is released under the Creative Commons NonCommercial license.

\noindent \textbf{QMSum}~\cite{zhong2021qmsum} consists of 1,808 query–summary pairs drawn from 232 meetings across multiple domains. We follow the topic-based partitioning provided by the dataset, focusing on Academic, Committee, and Product meetings. The training split contains 161 meetings and is used for model training, while the test split includes 35 meetings and is used for evaluation. The dataset is released under MIT License.

\noindent \textbf{ESConv}~\cite{liu2021towards} is an Emotional Support Conversation dataset collected in a help-seeker/supporter setting.
It contains 1,053 qualified multi-turn conversations with annotations on help-seeker emotions, problem types, feedback, emotion intensity, and supporter-side support strategies.
We use ESConv only as held-out evaluation for models trained on Empathetic Dialogues, testing cross-dataset transfer to unseen English emotional-support interactions.
The dataset is released under the Creative Commons Attribution-NonCommercial 4.0 International License.

\section{AI Assistants Usage}
\label{app:gpt}
AI assistants were used only as supporting tools during manuscript preparation, including for polishing prose, improving readability, reorganizing drafts, and assisting with  workflow automation. The core research ideas, experimental design, methodological decisions, and interpretation of results were developed and carried out by the authors. The authors carefully reviewed, edited, and verified all AI-assisted text and workflow outputs to ensure the accuracy, validity, and final presentation of the work.

\begin{table*}[htbp]
\centering
\renewcommand{\arraystretch}{1.2}
    \begin{tabularx}{\linewidth}{@{} >{\RaggedRight\arraybackslash}p{2.7cm} >{\RaggedRight\arraybackslash}X @{}}
    \toprule
    \textbf{Rubric Name} & \textbf{Description} \\
    \midrule
    \textbf{Emotional \newline Expressiveness} & Evaluates whether the emotional intensity of the response is appropriate for both the specific persona configuration and the immediate conversational context \cite{cuadra2024illusion, zhou_design_2020}. \\
    \midrule
    \textbf{Empathetic \newline Support \newline Quality} & Empathetic understanding following the dimensions in \citeauthor{sharma_computational_2020}: appropriate 
    \textit{Emotional Reactions}, accurate \textit{Interpretations} of user feelings, and supportive \textit{Explorations} via context-specific questioning \cite{cuadra2024illusion, rashkin2019towards, liu2021towards}. \\
    \midrule
    \textbf{\multirow{3}{*}{Consistency}} & Adherence to predefined persona attributes, including specific tone, vocabulary, values, and behavioral style as individual agents~\cite{zhang_personalizing_2018, higashinaka2018role}. \\
    \midrule
    \textbf{Response \newline Relevance} & Measures the logical consistency and contextual grounding of the response with respect to the immediate dialogue history and preceding turns \cite{liu_how_2016}. \\
    \midrule 
    \textbf{Social \newline Contribution} & Evaluates if the response meaningfully advances the interaction by adding relevant perspective, reassurance, or guidance, rather than offering generic acknowledgments \cite{liu_how_2016, zhou_design_2020, cuadra2024illusion}. \\
    \bottomrule
\end{tabularx}
\caption{Agent-specific criteria $\mathcal{C}$ for evaluating the overall discourse and agent coordination at the individual agent level.}
\label{tab:individual_criteria}
\end{table*}

\begin{table*}[htbp]
\centering
\renewcommand{\arraystretch}{1.2}
    \begin{tabularx}{\linewidth}{@{} >{\RaggedRight\arraybackslash}p{2.7cm} >{\RaggedRight\arraybackslash}X @{}}
    \toprule
    \textbf{Rubric Name} & \textbf{Description} \\
    \midrule
    \textbf{Engagement \& Contribution} & Proactive advancement of the discussion through novel viewpoints, synthesis of points, or facilitation of consensus~\cite{hong2024metagpt}. \\
    \midrule
    \textbf{Originality \& Specificity} & Use of original, contextually grounded insights from the interaction context rather than information that is generic or redundant with other agents~\cite{chang2024survey}. \\
    \midrule
    \textbf{Persona Fidelity} & Consistency with assigned roles, linguistic styles, and domain expertise in collaborative settings while avoiding generic AI-assistant behavior~\cite{wang2024rolellm}. \\
    \midrule
    \textbf{Relevance \& Coherence} & Logical alignment with preceding turns and overall conversation objectives to ensure semantic consistency and natural transitions~\cite{chan2024chateval}. \\
    \bottomrule
\end{tabularx}
\caption{Collective Interaction criteria $\hat{\mathcal{C}}$ for evaluating the overall discourse and agent coordination at the group level.
}
\label{tab:group_criteria}
\end{table*}

%

\begin{table*}[ht!]
\centering
\small
\begin{tabularx}{\textwidth}{@{} l l X X @{}}
\toprule
\textbf{Dataset} & \textbf{Agent} & \textbf{Description} & \textbf{Persona Characteristics} \\ \midrule
\textbf{ED} & Anchor & Specializes in high-vulnerability emotional support. The primary goal is to provide a safe space for users experiencing intense negative emotions like sadness, loneliness, or devastation. & validating, patient, reflective, emotionally-attuned, non-judgmental \\
 & Catalyst & Helps users navigate complex or frustrating situations (e.g., anger, anxiety, or guilt) and focus on the 'why' and 'what now' of a situation. & analytical, curious, solution-focused, socratic, perspective-shifting \\
 & Beacon & Amplifies the user's moments of joy, pride, and gratitude and specialize in 'capitalizing'—the process of expanding the benefits of positive news. & enthusiastic, strengths-focused, celebratory, energetic, positive-reinforcing \\ \midrule
\textbf{QMSum} & Minutes Scribe & Distill long meeting transcripts into concise summaries. & Objective, analytical, and highly structured. \\
 & Decision Logger & Given the meeting transcript, discuss the decisions made in the meeting. & analytical, evidence-based, systematic \\
 & Action Item Captain & Given the meeting transcript, discuss the actionable items so far for the meeting. & practical, action-oriented, precise \\
 & Critic & Critic for a workplace meeting. Review the meeting records from a critical perspective, discuss gaps, missing information, or overlooked issues. & critical, constructive, systematic \\ \bottomrule
\end{tabularx}
\caption{Definition of Agent Personas for Experiment Setup for datasets Empathetic Dialogues (ED) and QMSum.}
\label{tab:persona-definitions}
\end{table*}

\end{document}